\documentclass[sigconf]{acmart}
\usepackage{algorithm}
\usepackage{algpseudocode}
\usepackage{enumitem}
\AtBeginDocument{%
  }
\usepackage{listings}
\usepackage{xcolor}

\setcopyright{acmlicensed}
\copyrightyear{2018}
\acmYear{2018}
\acmDOI{XXXXXXX.XXXXXXX}

\acmConference[Conference acronym 'XX]{Make sure to enter the correct
  conference title from your rights confirmation emai}{June 03--05,
  2018}{Woodstock, NY}
\acmISBN{978-1-4503-XXXX-X/18/06}

\begin{document}
\lstset{
    escapeinside={(*@}{@*)},
    basicstyle=\ttfamily,   
    keywordstyle=\color{blue}, 
    commentstyle=\color{green}, 
    stringstyle==\color{red}, , 
    numbers=none,      
    backgroundcolor=\color{gray!10}, 
    showspaces=false,      
    showstringspaces=false, 
    showtabs=false,        
    tabsize=2,             
    breaklines=true
}
\renewcommand{\algorithmicrequire}{\textbf{Input:}}

\renewcommand{\algorithmicensure}{\textbf{Output:}}
\title{OpenSearch-SQL: Enhancing Text-to-SQL with Dynamic Few-shot and Consistency Alignment}

\author{Xiangjin Xie}
\email{xiexiangjin.xxj@alibaba-inc.com}
\affiliation{%
  \institution{Alibaba Cloud}
  \country{China}
}

\author{Guangwei Xu}
\email{kunka.xgw@alibaba-inc.com}
\affiliation{%
  \institution{Alibaba Cloud}
  \country{China}
}

\author{LingYan Zhao}
\email{zhaolingyan.zly@alibaba-inc.com}
\affiliation{%
  \institution{Alibaba Cloud}
  \country{China}
}

\author{Ruijie Guo}
\email{ruijie.guo@alibaba-inc.com}

\affiliation{%
  \institution{Alibaba Cloud}
  \country{China}
}

\renewcommand{\shortauthors}{Xiangjin et al.}

%
\begin{abstract}
Although multi-agent collaborative Large Language Models (LLMs) have achieved significant breakthroughs in the Text-to-SQL task, their performance is still constrained by various factors. These factors include the incompleteness of the framework, failure to follow instructions, and model hallucination problems. To address these problems, we propose OpenSearch-SQL, which divides the Text-to-SQL task into four main modules: Preprocessing, Extraction, Generation, and Refinement, along with an Alignment module based on a consistency alignment mechanism. This architecture aligns the inputs and outputs of agents through the Alignment module, reducing failures in instruction following and hallucination. Additionally, we designed an intermediate language called SQL-Like and optimized the structured CoT based on SQL-Like. Meanwhile, we developed a dynamic few-shot strategy in the form of self-taught Query-CoT-SQL. These methods have significantly improved the performance of LLMs in the Text-to-SQL task.

In terms of model selection, we directly applied the base LLMs without any post-training, thereby simplifying the task chain and enhancing the framework's portability. Experimental results show that OpenSearch-SQL achieves an execution accuracy(EX) of 69.3\% on the BIRD development set, 72.28\% on the test set, and a reward-based validity efficiency score (R-VES) of 69.36\%, with all three metrics ranking first at the time of submission. These results demonstrate the comprehensive advantages of the proposed method in both effectiveness and efficiency.
\end{abstract}




\keywords{Text-to-SQL, Agent, Large Language Model}

\received{20 February 2007}
\received[revised]{12 March 2009}
\received[accepted]{5 June 2009}

\maketitle

\section{Introduction}
\label{sec:intro}

Text-to-SQL task attempts to automatically generate Structured Query Language (SQL) queries from Natural Language Queries (NLQ). This task can improve the ability to access databases without the need for knowledge of SQL \cite{katsogiannis2023survey}.  
As the text-to-SQL problem is notoriously hard, these systems have been the holy grail of the database community for several decades \cite{katsogiannis2023survey}. Early work defined query answers as graph structures \cite{earlytxt2sql1,earlytxt2sql2} or based on syntactic structures for parsing the questions \cite{parse_survey1,parse_survey2}. Subsequent approaches have treated the Text-to-SQL task as a neural machine translation (NMT) problem \cite{rat, nmt1}. Recently, with the development of large language models, researchers have increasingly accomplished the task through methods such as the use of supervised fine-tuning (SFT) \cite{dubo,codes,Distilgpt4}, Chain-of-Thought(CoT) \cite{cot}, Agents \cite{dinsql,c3}, and In-Context Learning \cite{macsql,DAILsQL}, achieving results that greatly surpass previous methods.



\textbf{Limitation.} Although the methods driven by LLMs have significantly raised the upper limits of the Text-to-SQL task capabilities, our analysis of previous work reveals that:
\begin{enumerate}[label=\textbf{L\arabic*}]
    \item Due to the ambiguity of the overarching framework, there are some gaps at the methodological level. This has prevented these methods from reaching their potential. For example, there is a lack of verification of the stored information in the database, no error correction for the generated results, and issues related to the absence of few-shot learning.\label{l1}
    \item LLM-driven approaches often rely on multi-agent collaboration. However, due to the instability of LLMs and the lack of guaranteed coherence and coupling between agents, later-executing agents may not use or only partially use the outputs of previously running agents. This leads to accumulated hallucinations and performance loss.\label{l2}
    \item The instructions and steps that guide the LLMs significantly affect the quality of the generated SQLs. In the pre-LLM era, methods employing intermediate languages to generate SQL were developed to address this issue \cite{katsogiannis2023survey}, but current research on instruction construction remains insufficient.\label{l3}
\end{enumerate}


Inspired by the aforementioned challenges, we conducted a detailed analysis of the human workflow in completing the Text-to-SQL task: understanding the database structure and selecting the specific tables, columns, and values needed to get the SQL. This process often requires executing several simple SQL queries, such as investigating the specific forms of values for a certain column or retrieving how a particular value is stored in the database. Next, based on the specifics of the NLQ, the appropriate aggregation functions and SQL syntax are chosen to construct the backbone of the SQL query, followed by filling in the relevant statements. Finally, humans typically modify the SQL incrementally based on the query results until the requirements are met. For more complex questions, humans often choose to consult or borrow from others' formulations to seek solutions. Based on this approach, we developed the OpenSearch-SQL method, with detailed as follows.


\paragraph{\textbf{Framework.}} To address \textbf{\ref{l1}}, our research defines a \textbf{Standard Text-to-SQL framework} based on the process humans use to complete SQL construction. This can cover current research on LLM-driven Text-to-SQL tasks. We believe that a complete Text-to-SQL task framework should include four steps: \textbf{Preprocessing}, \textbf{Extraction}, \textbf{Generation}, \textbf{Refine}:
\begin{figure*}
    \centering
    \includegraphics[width=\textwidth]{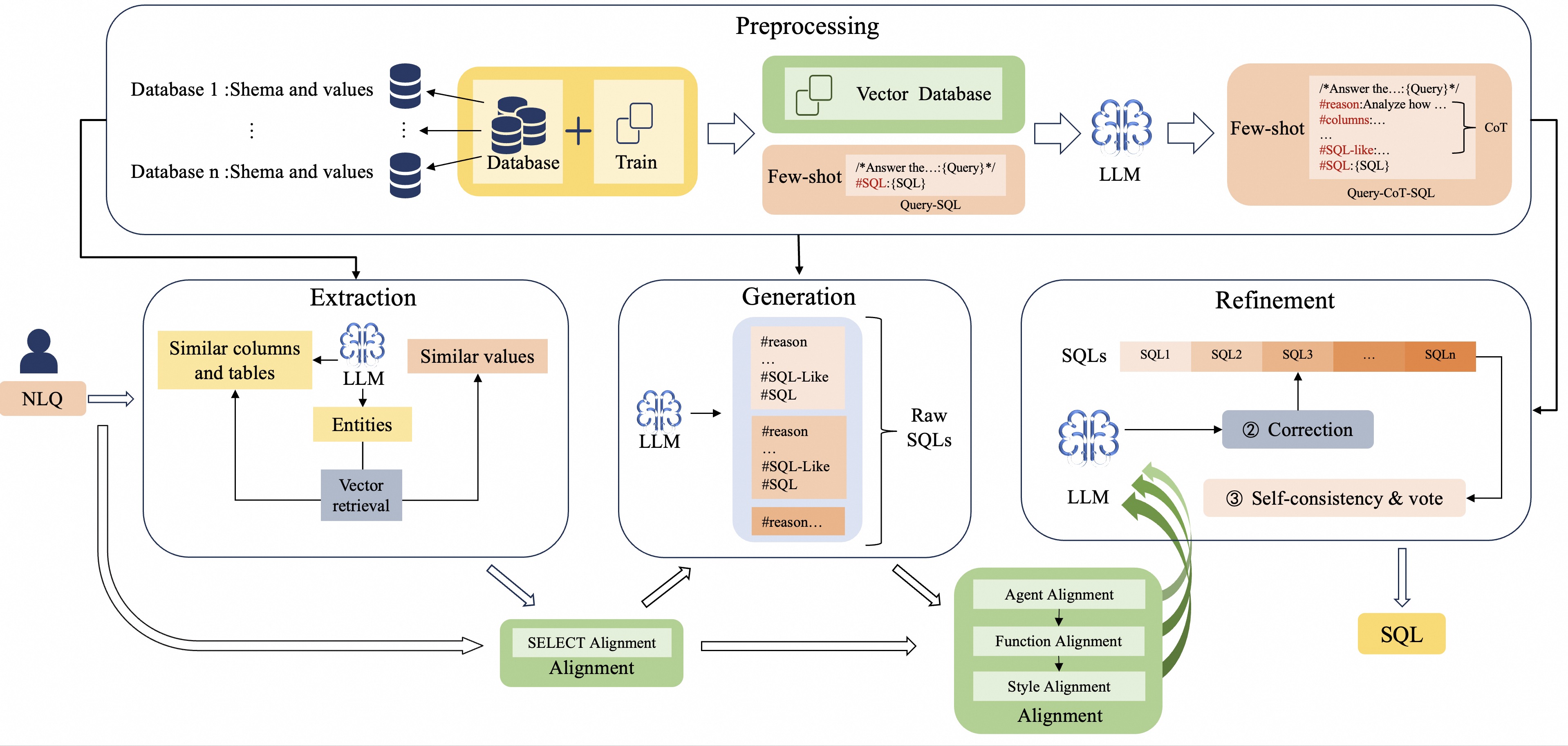}
    \caption{The basic framework of OpenSearch-SQL optimizes the Text-to-SQL task through a multi-agent collaboration approach based on consistency alignment.}
    \label{overview}
\end{figure*}
\begin{itemize}

\item \textbf{Preprocessing.} Processing and constructing all auxiliary information that is insensitive to the NLQs, including database schema information, embedding databases, few-shot libraries, etc.
  \item \textbf{Extraction.} Extract and select the necessary elements for generating SQL, including sub NLQ, few-shot, as well as tables, columns, values, etc., from the database.
  \item \textbf{Generation.} Translate NLQ into SQL based on the SQL syntax, few-shot and the prepared information. 
  \item \textbf{Refine.} Check and optimize the SQL based on the execution results using alignment strategies and rules. Then, select the final SQL based on self-consistency and voting results.
\end{itemize}

For \textbf{\ref{l2}}, we analyzed common hallucinations in multi-agent collaboration-driven Text-to-SQL tasks and designed a consistency alignment-based \textbf{ alignment agent} for these hallucinations. By ensuring consistency in the functions, inputs, and outputs of specific agents, our aim is to reduce hallucinations during the model generation process. This alignment approach can also be extended to make targeted adjustments to the agents' outputs. As shown in Figure \ref{overview}, we proposed the OpenSearch-SQL framework, which consists of a standard Text-to-SQL framework combined with alignments.

Furthermore, we extended our research on multi-agent methods driven by LLMs using this framework. For \textbf{\ref{l3}}, we developed a self-taught few-shot augmentation mechanism that supplements query-SQL pairs with \textbf{Chain-of-Thought(CoT)} information, creating \textbf{Query-CoT-SQL} pairs. This enriched few-shot approach can enhance the performance of LLMs. Additionally, we designed an intermediate language named SQL-Like within the CoT logic, allowing LLMs to first generate the backbone of the SQL before the specific SQL details, thus reducing the complexity of the task for the models.



In real-world applications, preparing a large amount of customized SQL training data for specific query tasks is not an easy task. Therefore, considering the needs of the real world, our goal is not only to achieve good results in specific evaluation tasks, but also to ensure that the proposed method has strong transferability. We aim for it to be easily applicable to various database query tasks without the need for post-training. This means that when faced with any issue, we can simply modify the agent and alignment methods without the need to retrain the model, thus meeting user needs with high quality.

Based on this, we directly implemented our approach on a pre-trained LLM model without employing any supervised fine-tuning (SFT) or reinforcement learning (RL). In terms of experiments, we chose the widely recognized BIRD dataset \cite{Bird} for testing, which records three metrics: Execution Accuracy(EX) on the development set, EX on the test set, and the Reward-based Valid Efficiency Score(R-VES) on the test set. Our experimental results indicate that we achieved 69. 3\% EX on the Dev set, 72. 28\% EX on the Test set, and an R-VES score of 69.3\%. All three metrics ranked first on the leaderboard at the time of submission, demonstrating that our method not only generates accurate SQL statements but also has a clear advantage in terms of time efficiency. We hope that this consistency alignment-based approach will inspire future research to explore more efficient consistency alignment strategies.

\textbf{Contribution} In summary, our contributions are as follows.
\begin{itemize}

\item To the best of our knowledge, we are the first to propose a multi-agent cooperative framework based on consistency alignment for the Text-to-SQL task. This alignment mechanism significantly reduces the loss of information transmission between Agents and the hallucinations in the results generated by the Agents.

\item We introduce a novel self-taught dynamic few-shot method and a SQL-Like CoT mechanism. This enhances the performance and stability of LLMs when generating SQL.

\item Our proposed method, OpenSearch-SQL, did not utilize fine-tuning or introduce any specialized datasets. At the time of submission, it achieved 69.3\% EX on the BIRD benchmark validation set, 72.28\% EX on the test set, and R-VES of 69.36\%, all ranking first.
\end{itemize}

\section{Preliminary}

We first introduce the two important modules this paper relies on: Large Language Models (LLMs) and the Text-to-SQL task.
\subsection{Large Language Models} 

Pre-trained large language models have acquired extensive language knowledge and semantic information through unsupervised learning on large-scale textual data. This pre-training typically employs autoregressive architectures (such as GPT-like models) or autoencoding structures (such as BERT-like models). The LLMs can achieve excellent performance on various downstream tasks through fine-tuning. Since the advent of the deep learning era for large language models, there have been significant advancements in Text-to-SQL tasks compared to classical models and methods, with current SOTA-level work fully relying on the performance of LLMs.

\subsection{Text-to-SQL}
The Text-to-SQL task can be defined as: a task that translates a NLQ $\mathit{Q}$ into an SQL query $\mathit{Y}$ based on database information $\mathit{S}$, a specific model $F(\cdot | \theta)$, and a particular prompt $\mathit{P}$. The information in the database is defined as $\mathit{S}$. Therefore, the Text-to-SQL task can be defined as: 
\begin{equation}
\mathit{Y}=F(\mathit{Q},\mathit{S},\mathit{P} | \theta),
\end{equation}




In the current state-of-the-art Text-to-SQL tasks, multi-agent collaboration has become a key strategy for enhancing performance. Common modules include value retrieval, schema linking, few-shot driving, CoT prompting, SQL correction, and self-consistency. Inspired by these methods, this paper first systematically organizes these modules. We categorize existing methods into four stages: \textbf{Preprocessing}, \textbf{Extraction},\textbf{ Generation}, and \textbf{Refinement}. The \textbf{preprocessing} is independent of a specific NLQ and aims to obtain clear information about the database structure and prepare other auxiliary information that aids the model in generating SQL; the \textbf{Extraction} analyzes specific queries and filters out high-quality auxiliary information from the prepared information; the\textbf{ Generation} organizes the extracted information and generates candidate SQLs; and the refinement stage further optimizes the candidate SQLs to enhance the quality of SQLs.

Based on this foundation, we propose OpenSearch-SQL, a Text-to-SQL framework based on \textbf{dynamic few-shot} and \textbf{consistency alignment} mechanisms. The design of dynamic few-shot aims to strengthen the performance of each module across the various stages of the task, while the \textbf{Alignment} module is used to align the different stages of Text-to-SQL, thereby reducing the hallucination produced by LLMs in multi-agent collaboration workflows. The design of OpenSearch-SQL simplifies the complex instruction system, focusing on improving multi-agent collaboration effectiveness through alignment. In the following sections, we will provide a detailed introduction to the specific components of the OpenSearch-SQL framework.

\subsection{Hallucination}
Hallucination \cite{hallucination} is a common problem in deep learning, usually referring to situations where the content generated by a model does not align with reality or expected outcomes, i.e., the model produces inaccurate or unreasonable information. Textual hallucinations can manifest in the following forms:


\begin{itemize}
    \item Irrelevant content: The text generated by the model is unrelated to the input or context, or it includes irrelevant details.
    \item Incorrect facts: The generated text contains erroneous facts or information that do not exist or are inaccurate in the real world.
    \item Logical errors: The generated text is logically inconsistent or unreasonable.
\end{itemize}

In Text-to-SQL, this manifests as: results containing non-existent database information, failure to follow instructions in the prompt, and typographical and syntactic errors. Additionally, for convenience in handling, in the Text-to-SQL task, we also include the result deviations caused by randomness at non-zero temperatures and errors due to minor prompt changes.
Because the essence of hallucination is the discrepancy between the training process and the actual use scenarios of the mode. Therefore, for hallucination problems, common handling methods include: post-training specifically for hallucination problems, using the Self-Consistency mechanism to reduce randomness, and post-processing. Since this paper focuses on optimizations at the architectural level, our work aims to reduce hallucinations through the last two methods mentioned above.

\label{sec:method}
\section{Methodology}
In this section, we detail the specifics of OpenSearch-SQL. We provide a more in-depth description of the core concepts of Alignments and dynamic few-shot before outlining the entire process. To ensure a smooth explanation, we integrate each stage with its corresponding Alignments and describe them in the order they operate within the OpenSearch-SQL framework.

\subsection{Alignments}

LLM hallucination \cite{hallucination} is a critical problem affecting the usability of LLMs, and it is also present in the Text-to-SQL task. Furthermore, the cumulative errors that arise from multi-agent collaboration exacerbate the hallucination problem in LLMs. For example, if an agent responsible for the \textbf{Extraction} function selects columns from the database in a certain way and generates incorrect column names due to model hallucination, this error will persist in the subsequent \textbf{Generation}, as it lacks knowledge of the correct database structure. Consequently, the hallucination produced by the previous agent is inherited, making it difficult for hallucinations to spontaneously disappear in a multi-agent LLM workflow; the total amount of hallucination is almost monotonically non-decreasing.

Based on this phenomenon, we define the Alignment Agent as :
\begin{equation}
A_{Aligment}(x+A'(x))=A(x)-A'(x),
\end{equation}
The input for the agent is the input $x$ for the agent to be aligned, along with $A'(x)$. The output is $A_{Aligment}(x+A'(x))$, which is obtained through optimization by a large model and rules. Empirical observations suggest that the difference between $A'(x)$ and $A(x)$ is mainly due to hallucinations in the model. Therefore, the function of Alignment is to align the current Agent's input information with its output result, so that this alignment can reduce the hallucinations between the expected output and the actual output.



In the Text-to-SQL task, common hallucination phenomena mainly refer to failures in instruction following and instability in output results. This is specifically manifested as: generating non-existent columns, changing database column names, syntax errors, mismatching database values with columns, and failing to adhere to the rules set in the prompt. To address these problems, we propose a consistency alignment method: after each agent completes its output, an \textbf{Alignment} agent is used to align the current agent's output with that of the upstream agent, ensuring that the functionalities of the various agents achieve logical consistency, and the aligned result is passed on to the downstream agent. This mechanism is somewhat analogous to residual connections \cite{resnet}, effectively extending the chain of collaboration among multiple agents while minimizing the introduction of hallucinations. The details of each Alignment will be presented in detail in the following Agent introduction.

\subsection{Self-Taught Fewshot}\label{dynamic fewshot}


Few-shot is an important method to assist LLMs in generation, MCS-SQL\cite{mcssql}, DAIL-SQL\cite{DAILsQL} studied the critical role of problem representation in the Text-to-SQL task and proposed using question similarity to select appropriate few-shots to drive LLMs in generating SQL, achieving notable results. 

Inspired by this, we attempted to use dynamic few-shots to enhance the efficiency of agents in the Text-to-SQL task. Additionally, we considered how to better leverage few-shots by extracting more information from the samples. Therefore, in OpenSearch-SQL, we first used \textbf{Masked Question Similarity (MQs)} \cite{mqs} to select similar queries. Then, we upgraded the few-shots in the Query-SQL format by self-taught. For the Query-SQL pairs shown in Listing \ref{query-SQL}, we used an LLM to supplement the CoT information to transform the NLQ into SQL. As shown in Listing \ref{query-CoT-SQL}, this resulted in \textbf{Query-CoT-SQL} pairs containing logical information. Compared to simple Query-SQL pairs, these self-taught few-shots provide richer information.

Then, for the error correction of the \textbf{Refinement}, we prepared different few-shots for various error types, enabling LLMs to more clearly understand how to correct SQL based on different errors during \textbf{Refinement}. The format of the few-shots is shown in Listing \ref{reffewshot}, the specific errors in the \textbf{Raw SQL} and the \textbf{Advice of correction} correspond to the error type.
\begin{lstlisting}[caption={Format of Few-shot in Query-SQL Pair},label ={query-SQL}]
/* Answer the following:(*@\color{red}{\{question\}}@*) */
#SQL: (*@\color{red}{\{SQL\}}@*)
\end{lstlisting}

\begin{lstlisting}[caption={Format of Few-shot in Query-CoT-SQL Pair},label ={query-CoT-SQL}]
/* Answer the following:(*@\color{red}{\{question\}}@*) */
#reason: Analyze how to generate SQL based on the question.
#columns: All columns ultimately used in SQL
#values: the filter in SQL
#SELECT: SELECT content table.column.
#SQL-like: SQL-like statements ignoring Join conditions
#SQL: (*@\color{red}{\{SQL\}}@*)
\end{lstlisting}

\begin{lstlisting}[caption={Format of Few-shot in Correction},label ={reffewshot}]
{"Result: None": """/* Fix the SQL and answer the question */
#question: (*@\color{red}{\{question\}}@*)
#Error SQL: (*@\color{red}{\{Raw SQL\}}@*)
Error: Result: None
#values: (*@\color{red}{\{values in Database\}}@*)
#Change Ambiguity: (*@\color{red}{\{Advice of correction\}}@*)
#SQL:(*@\color{red}{\{corrected SQL\}}@*)""",
...}
\end{lstlisting}

\subsection{Preprocessing}
In the \textbf{Preprocessing}, we constructed the database based on its true structure. Additionally, to ensure that the SQL generated by LLMs aligns with the actual state of the database, we indexed the values within the database. This allows the SQL to avoid errors that might arise from small character discrepancies. It's worth noting that we index only string-type data to save the space required for building the retrieval database.

In addition, the construction of dynamic Fewshot has also been completed in this phase. This includes the addition of CoT information and correction few-shot examples designed for different error types.

Overall, during the preprocessing stage, the inputs are the information from the database and the training set. The outputs are a vector database, Few-shot in the form of Query-CoT-SQL, and the database schema. This process is automated, does not require human intervention, and is entirely driven by an Agent.

\subsection{Extraction}


The goal of \textbf{Extraction} is to prepare the necessary information based on specific NLQs, including schema linking, stored database values, few-shot examples, and any required instructions. This part is decoupled from the specific database query language and is only related to the format of data storage and the auxiliary information needed to solve the problem.
For instance, C3-SQL\cite{c3} utilizes Clear Prompting (CP) to provide effective prompts, while DIN-SQL\cite{dinsql} employs schema linking and classification \& decomposition to categorize and break down the NLQ.


\begin{lstlisting}[caption={Format of Extraction},label ={extract_prompt}]
(*@\colorbox{yellow}{Input:}@*)
/* Database schema */
(*@\color{red}{\{db\_info\}}@*)
(*@\color{red}{\{rule\}}@*)
/* Answer the following: (*@\color{red}{\{query\}}@*) */

(*@\colorbox{yellow}{Output:}@*)
(*@\color{red}{\{reason\}}@*)
(*@\color{red}{\{column\}}@*)
(*@\color{red}{\{values\}}@*)
\end{lstlisting}

The \textbf{Extraction} process in OpenSearch-SQL consists of entity extraction, value extraction, and column filtering. After \textbf{Extraction,} we use Info Alignment to align the extracted information with the input to produce the final output. We present the format of the extraction prompt and its input and output in the listing \ref{extract_prompt}. We obtain candidate columns and values through LLM and directly extract entities from the NLQ using another simple prompt. The details are outlined below:

\paragraph{\textbf{Entity Extraction}}
To find similar values in the database and perform column filtering, we first use the LLM understand and process the NLQ and basic database information to extract potential entities. Then, we organize some predefined entity terms together with these entities to perform the subsequent values retrieval and column filtering. 

\paragraph{\textbf{Values Retrieval.}}
From the extracted entities, we perform vector retrieval to find results with the highest embedding similarity to the entity words. Due to the nature of embedding similarity, this method can prevent mismatches caused by typos or other character-level differences. Additionally, for phrases and longer texts, we perform split retrieval to avoid recall failures due to differences in database storage formats. During the recall phase, we filter out portions of the top K most similar entities that fall below a specific threshold, retaining them as the final results.


\paragraph{\textbf{Column Filtering.}}
In detail, we utilize two methods to recall relevant tables and columns. First, we employ a large language model (LLM) to select tables and columns related to the natural language query (NLQ) from complete database information. Then, we use vector retrieval to find columns in the database with similarity to NLQ entities exceeding a certain threshold, integrating these to form a preliminary subset of the final Schema information. Although this multi-path recall method may lack some precision in filtering, it is lighter and more streamlined in process.

\paragraph{\textbf{Info Alignment}} At the end of the extraction phase, we use an Info aligment agent to align the style of the SELECT statement: extracting phrases or clauses from the NLQ that correspond one-to-one with the generated SELECT content. This ensures that both the quantity and order of the SELECT content meet expectations. To avoid misselection caused by columns with the same name in different tables, we expand the schema information by reintegrating the primary keys of each table and all columns with the same name as the columns selected in column filtering into the column filtering results, ensuring the accuracy and comprehen siveness of the selected columns. Finally, all the extracted information and the content of the Alignment will be input into the \textbf{Generation}.

\subsection{Generation}
The definition of \textbf{Generation} is: using appropriate methods to drive LLMs to generate specific components and complete SQL. MAC-SQL \cite{macsql} generates sub-SQLs and then assembles them into the final result, while DAIL-SQL \cite{DAILsQL} drives LLMs to generate SQL through few-shot learning. Additionally, some methods \cite{chess, Distilgpt4, mcssql} utilize LLMs during Supervised Fine-Tuning (SFT) or generate multiple sets of SQL to complete tasks.
\definecolor{darkgreen}{rgb}{0.0,0.5,0.0}
\begin{lstlisting}[caption={Example of Generation},label ={generate}]
(*@\colorbox{yellow}{Input:}@*)
(*@\color{red}{\{fewshot\}}@*)
/* Database schema */
(*@\color{red}{\{db\_info\}}@*)
(*@\color{red}{\{similar values\}}@*)
(*@\color{red}{\{rule\}}@*)
(*@\color{darkgreen}{Rule Example: For parts involving division that contain integer types, CAST them to REAL}@*)

/* Answer the following: How many patients with a normal Ig A level came to the hospital after 1990? normal Ig A level refers to IGA > 80 AND IGA < 500; */
(*@\color{red}{\{info alignment\}}@*)
(*@\color{darkgreen}{Info Alignment Example: SELECT content: [How many patients]}@*)

(*@\colorbox{yellow}{Output:}@*)
#reason: The question wants to count the number of patients with a normal Ig A level who came to the hospital after 1990, so SELECT will count distinct patients based on the specified conditions.
#columns: Patient.ID, Laboratory.IGA, Patient.`First Date`
#values: normal Ig A level refers to Laboratory.IGA > 80 AND Laboratory.IGA < 500; came to the hospital after 1990 refers to strftime('%Y', Patient.`First Date`) >= '1990'
#SELECT: How many patients refer to COUNT(DISTINCT Patient.ID)
#SQL-like: Show COUNT(DISTINCT Patient.ID) WHERE Laboratory.IGA > 80 AND Laboratory.IGA < 500 AND YEAR(Patient.`First Date`) >= '1990'
#SQL: SELECT COUNT(DISTINCT T1.ID) FROM Patient AS T1 INNER JOIN Laboratory AS T2 ON T1.ID = T2.ID WHERE T2.IGA > 80 AND T2.IGA < 500 AND strftime('%Y', T1.`First Date`) >= '1990'
\end{lstlisting}



\paragraph{\textbf{SQL Generation.}} As shown in Listing \ref{generate}, we chose to use the progressive generation $+$ dynamic few-shot to better drive LLMs in generating SQL. Specifically, progressive generation is a customized CoT approach, the format is the same as Listing \ref{query-CoT-SQL}. We first defined SQL-Like: a type of SQL language that ignores specific syntactical elements (such as JOINs and the formatting of functions), aiming to encourage LLMs to focus more on the logic within SQL rather than the formatting. We then instruct the model to sequentially provide the \textbf{analysis}, the contents of the \textbf{SELECT statement}, \textbf{relevant columns} and \textbf{values}, as well as the \textbf{SQL-Like} and \textbf{SQL} outputs. This progressive generation emphasizes the LLMs' understanding of the structure of the generated SQL, while also faciliting the identification of errors in the LLMs' reasoning. For individual NLQs, we allow LLMs to generate multiple candidate SQL queries. For the construction of few-shot examples, the system retrieves the top $K_f$ most similar queries based on MQs, and uses their corresponding Query-CoT-SQL forms as the few-shot examples in the final instruction.

Listing \ref{generate} demonstrates the composition of inputs and outputs during the generation phase through an example.
\begin{enumerate}
    \item In \textbf{input}, we provide the necessary output requirements and database information by the Rules and Database schema, and we offer similar examples through few-shot in Query-CoT-SQL format. \textbf{Similar Values} provide specific values from the database that are relevant to the NLQ, while \textbf{Info Alignment} aligns the SELECT style of the dataset with the LLMs.
    \item In the \textbf{output}, we require the following sequential generation: \textbf{Reason}: Analysis of NLQ, \textbf{Columns}: The relevant columns in the SQL, \textbf{Values}: The related values, \textbf{SELECT}: The content of the SELECT statement. Then, generate \textbf{SQL-Like} query that ignores syntax formatting and ultimately produces the final SQL queries.
\end{enumerate}

\begin{lstlisting}[caption={Examples of Alignments},label ={align}]
(*@\color{blue}{\textbf{Agent Alignment:}}@*)
(*@\color{darkgreen}{Raw SQL}@*): SELECT ID FROM table WHERE table.name= 'John'
value in Database: table.name='JOHN'
(*@\color{darkgreen}{Aligned SQL}@*): SELECT ID FROM table WHERE table.name= 'JOHN'

(*@\color{blue}{\textbf{Function Alignment:}}@*)
(*@\color{darkgreen}{Raw SQL}@*): SELECT ID FROM table ORDER BY MAX(score)
(*@\color{darkgreen}{Aligned SQL}@*): SELECT ID FROM table GROUP BY ID ORDER BY score

(*@\color{blue}{\textbf{Style Alignment:}}@*)
(*@\color{darkgreen}{Raw SQL}@*): SELECT ID FROM table ORDER BY score DESC LIMIT 1
(*@\color{darkgreen}{Aligned SQL}@*): SELECT ID FROM table WHERE score IS NOT NULL ORDER BY score DESC LIMIT 1


\end{lstlisting}
\paragraph{\textbf{Alignments}} 
The errors in the SQL query generation process primarily arise from differences in syntax, databases, natural language queries (NLQ), and dataset styles. Additionally, employing large models to sample at higher temperatures for diverse answers may introduce errors and noise. Therefore, we attempt to reduce the biases caused by these discrepancies through an alignment mechanism.
Specifically, as shown in the listing \ref{align}, the alignment mechanism consists of the following three components:
\begin{itemize}
    \item Agent Alignment: Ensures that the columns and values from the database are correctly represented in the SQL. If there is a mismatch, corrections are made. A common example is a WHERE condition in SQL that does not match the information stored in the database.
    \item Function Alignment: Standardizes the SQL aggregate functions to prevent errors caused by incorrect expressions. This includes handling inappropriate AGG functions, nestings, and redundant JOINs.
    \item Style Alignment: Addresses issues related to dataset characteristics, such as the use of IS NOT NULL and the choice between MAX and LIMIT 1.
\end{itemize}

After completing the above steps, the SQL is executed, and during the \textbf{Refinement}, it is optimized and the final selection is made based on the execution results.
\subsection{Refinement}

The work of \textbf{Refinement} involves optimizing and selecting the generated SQLs. This includes correcting errors based on execution results, and selecting the best answer from multiple candidate SQLs.

\begin{figure}
    \centering
    \includegraphics[width=1.0\linewidth]{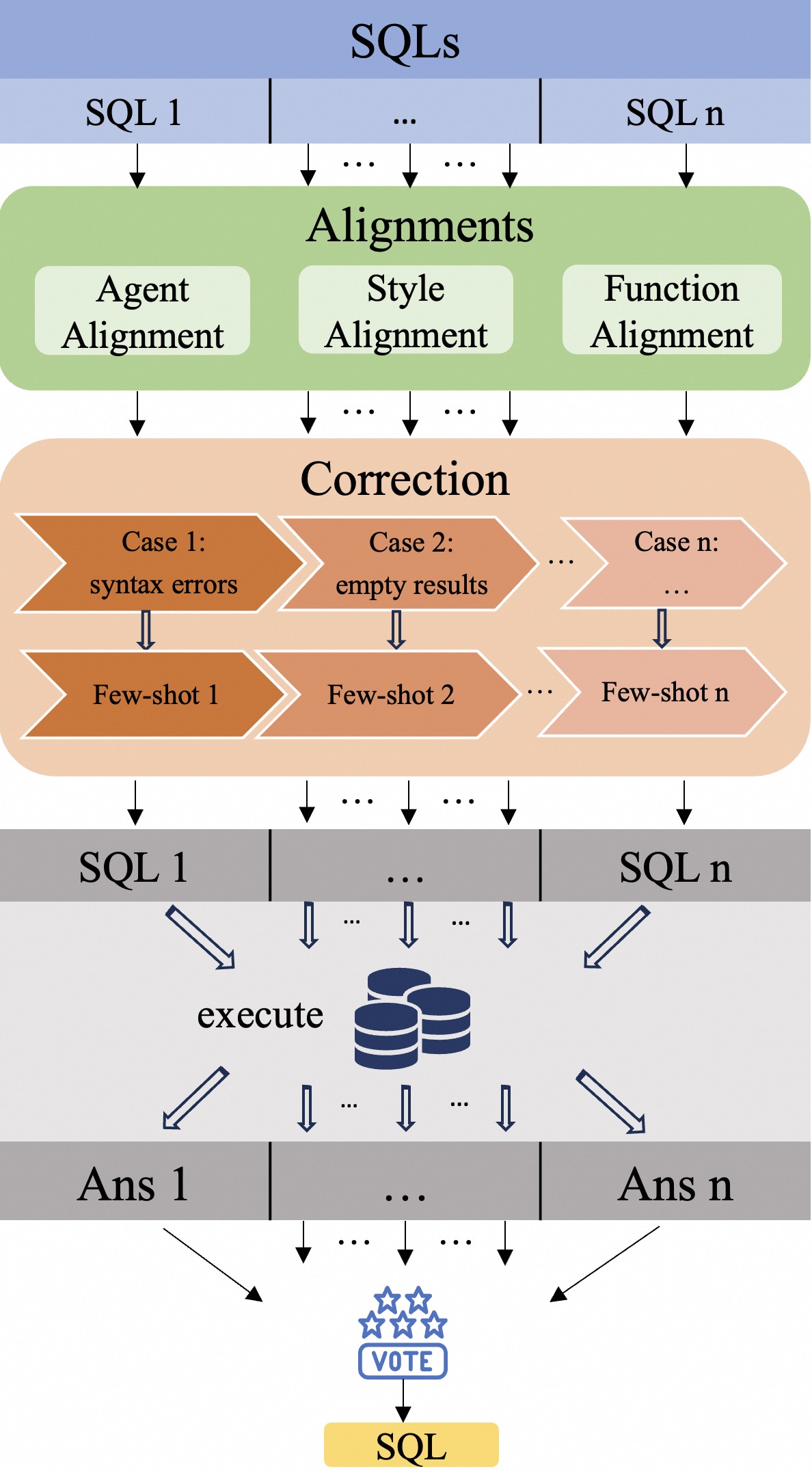}

    \caption{Architecture of the \textbf{Refinement}: Optimizing SQLs through Correction and Consistency \& Vote to get the best SQL.}\label{Refinement}
\end{figure}

Due to the variability of results generated by LLMs and the frequent issues with instruction adherence in complex tasks, we propose a consistency alignment mechanism based on checking to enhance the quality of SQL. As illustrated in Figure \ref{Refinement}, after aligning the generated SQLs, the refinement consists of the following two steps:
\begin{enumerate}
    \item \textbf{Correction}: Execute the SQL and fix it based on the error details in the execution results, such as syntax errors or empty results. Each type of error corresponds to different few-shot and error-correction instructions. The purpose of this step is to avoid issues such as lack of results or execution failures caused by minor details.
    \item \textbf{Self-consistency \& vote}: Exclude SQLs that cannot be fixed and those that result in empty answers and choose the option with the highest output consistency in the execution results. Additionally, among SQL queries with the same answers, we select the one with the shortest execution time. This approach achieves a dual enhancement in both execution results and execution speed for the final selected SQL. Given the set $K=\{R_1,R_2...R_n\}$, where $R_i=(SQL_i,ans_i,t_i)$. $SQL_i$ represents the SQL query, $ans_i$ represents the result of the execution, and $t_i$ represents the execution time. We obtain the final SQL query using:
\end{enumerate}

\begin{equation}
\begin{aligned}[t]
&\text{SELECT \textbf{SQL} FROM \textbf{K} WHERE \textbf{ans} =} \\
&\text{(SELECT \textbf{ans} FROM \textbf{K}} \\
&\text{GROUP BY \textbf{ans} ORDER BY} \\
&\text{COUNT(*) DESC LIMIT 1)} \\
&\text{ORDER BY \textbf{t} LIMIT 1}
\end{aligned}
\end{equation}

It is worth mentioning that self-consistency often leads to higher costs. In the evaluation, when only one SQL is generated in the \textbf{Generation} and the self-consistency mechanism is not used, OpenSearch-SQL, v2, still ranks first in the BIRD benchmark.

\subsection{Algorithm}
In this section, we provide an overall summary of OpenSearch-SQL and present the entire framework in Algorithm \ref{alg}. In this algorithm, we take an NLQ as an example, encompassing each specific step from preprocessing to the final generation of SQL.

It can be seen that we first process the database table by table, field by field, during preprocessing to build an index database $D_v$ of values and columns' names, while also generating the original schema $S$ of the database. Next, for the Query-SQL pairs in the training set, we use 
$M_L$ to generate CoT information to construct the new few-shot set $F$. In the main function, we first process the NLQ $Q$ using a LLM. Through $A_E$ we obtain the entity information $E_Q$ and select the column $t.c_Q$. Then, we retrieve similar values $v_Q$and expand the database information $t.c_E$. These details are integrated through Alignment $A_a$ to create a schema $S_Q$ tailored for $Q$. The next step is to obtain the few-shot $f_Q$ in Query-CoT-SQL format relevant to the question based on $MQs(Q)$. Next, SQLs are generated through $A_G$ and aligned using $A_a$ to obtain $SQLs_A$. Finally, execution errors are corrected with $A_R$, and the SQL with the highest consistency and shortest execution time, $SQL_R$, is selected as the final answer.

\begin{algorithm}
\caption{Algorithm of OpenSearch-SQL}\label{alg}
\begin{algorithmic}[1]
\Require Target database D, user question Q, train set T
\State  Extraction Agent $A_E$, Generation Agent $A_G$, Refinement Agent $A_R$, Alignment Agent $A_a$, vector model $M_E$, LLM $M_L$, Fewshot Set F, Database Schema S, vector database $D_v$
\State \Comment{Preprocessing}
\For{Table $t$ in $D$} \Comment{Build vector database}
    \For{column $c$, values $v$ in $t$}
        \State get description $d$
        \If{$v$ is String}
            \State $D_v \gets M_E(v), M_E(t.c) : t.c$ \Comment{Add index to $D_v$}
        \EndIf
        \State $S \gets S + d$
    \EndFor
\EndFor
\For{Query-SQL pair ($q_t, s_t$) in $T$} \Comment{Create fewshot}
    \State $CoT_t \gets M_L((q_t, s_t))$
    \State Add Query-CoT-SQL pair ($q_t, CoT_t, s_t$) to $F$
\EndFor
\State \Comment{Main Process}
\State $E_Q, t.c_Q \gets A_E(Q, S)$
\State $v_Q \gets D_v(E_Q)$ \Comment{Get similar values to $Q$}
\State $t.c_E \gets D_v(E_Q)$ \Comment{Get table, column related to $Q$}
\State $S_Q \gets A_a(Q, S, t.c_E, t.c_Q, v_Q)$ \Comment{New Schema}
\State $f_Q \gets F(MQs(Q))$
\State $SQLs \gets A_G(f_Q, S_Q, Q)$
\State $SQLs_A \gets A_a(SQLs, S_Q, Q)$ \Comment{Align SQLs to $Q$ and $S_Q$}
\State $SQL_R \gets A_R(SQLs_A, S_Q)$ \Comment{Refinement of SQLs}

\State \Return $SQL_R$
\end{algorithmic}
\end{algorithm}

\subsection{Optimization}


OpenSearch-SQL has achieved significant results, yet there is still plenty of room for optimization. Currently, we haven't fully focused on the detailed adjustments of prompts and the precise selection of columns and values in OpenSearch-SQL. Moreover, in the generation tasks, we only use a single prompt as the instruction for the LLM to generate a SQL candidate set, without additional optimization. On the other hand, the SFT model shows great potential in improving Text-to-SQL performance. Research such as CHESS \cite{chess}, MCS-SQL \cite{mcssql} and distillery \cite{Distilgpt4} has already demonstrated the effectiveness of these methods, suggesting further optimization can be pursued in these directions.

In terms of methods within OpenSearch-SQL, we believe that few-shot approaches are not limited to Query-CoT-SQL pairs; there are other options available, offering potential for enhanced performance. Additionally, the research on alignments is still in its early stages, indicating there is room for further development in improving Text-to-SQL task performance.

\section{Experiments}
\label{sec:experiments}
In this section, we demonstrate the effectiveness of OpenSearch-SQL through experiments and explore the role of its various module.
\subsection{Experimental Setup}
\paragraph{Datasets}
We show the characteristics of these two datasets in Table \ref{tabdataset}. Considering the specific queries and SQL complexity of the datasets. Bird has relatively fewer types of databases but with more complex database structures and a higher average difficulty of SQL. In contrast, Spider has a rich variety of databases, but the average difficulty of SQL is relatively lower. 

\paragraph{\textbf{BIRD}} \cite{Bird} (Big Bench for Large-scale Database Grounded Text-to-SQL Evaluation) represents a pioneering, cross-domain dataset that examines the impact of extensive database contents on text-to-SQL parsing. BIRD contains over 12,751 unique question-SQL pairs, 95 big databases with a total size of 33.4 GB. It covers more than 37 professional domains, such as blockchain, hockey, healthcare, and education, etc. 
The BIRD dataset released a cleaner version of the dev set on 4 July 2004. However, for the sake of fairness, we will still evaluate our results using the dev data from before July.

\paragraph{\textbf{Spider}}\cite{spider} is a large-scale complex and cross-domain semantic parsing and text-to-SQL dataset annotated by 11 Yale students. The goal of the Spider challenge is to develop natural language interfaces to cross-domain databases. It consists of 10,181 questions and 5,693 unique complex SQL queries on 200 databases with multiple tables covering 138 different domains. In Spider, we perform tuning on the dev set, assuming the test set is not accessible.


\begin{table}[t]
\centering
\renewcommand\arraystretch{2} 
\resizebox{\linewidth}{!}{
\begin{tabular}{c|ccccc}
\hline
Dataset  & train  & dev &test  & domains &databases\\ \hline \hline
Spider  & 8659 & 1034 &2147 & 138 &200 \\ \hline
Bird & 9428   & 1534 & 1789  & 37 &95\\ \hline
\end{tabular}}
\caption{Statistics of the datasets.}\label{tabdataset}
\end{table}

\paragraph{\textbf{Evaluation}}
Following the evaluation criteria from BIRD \cite{Bird} and the Spider test suite \cite{test-suite}, we assessed two metrics: \textbf{Execution Accuracy (EX)} and \textbf{Reward-based Valid Efficiency Score (R-VES)}. \textbf{EX} is defined as the proportion of identical execution results between the predicted SQL and the gold SQL. The purpose of \textbf{R-VES} is to measure the execution efficiency of the predicted SQL when performing the same task.

\paragraph{\textbf{Baselines}}
We selected the LLM-driven methods from the BIRD and Spider rankings as baselines:
\begin{enumerate}

\item \textbf{GPT-4} \cite{openai2024gpt4} employs a zero-shot text-to-SQL prompt for SQL generation.

\item \textbf{DIN-SQL} \cite{dinsql} divides questions into different types through multiple modules and instructs LLM to generate the final SQL through various prompts.

\item \textbf{DAIL-SQL} \cite{DAILsQL} assists LLMs in generating SQL by selecting similar question-SQL pairs for different questions as few-shot examples.

\item \textbf{MAC-SQL} \cite{macsql} simplifies the challenges LLMs face by using sub-databases and sub-questions to generate SQL.

\item \textbf{MCS-SQL} \cite{mcssql} generates multiple sets of SQL using multiple prompts and employs a unified Multiple-Choice Selection (MCS) to select the final SQL.

\item \textbf{C3-SQL} \cite{c3} has constructed a systematic Zero-shot Text-to-SQL approach through three modules: Clear Prompting (CP), Calibration with Hints (CH), and Consistent Output (CO).

\item \textbf{CHESS} \cite{chess} improves the effectiveness of SQL generation by constructing efficient retrieval and schema pruning methods to reduce the interference of redundant information.

\item \textbf{Distillery} \cite{Distilgpt4} suggests that the importance of schema linking has diminished with the development of LLMs, and the fine-tuned SFT GPT-4o has achieved optimal results.
\end{enumerate}

\paragraph{\textbf{Implementation Details}}
To demonstrate the capabilities of OpenSearch-SQL v2, the specific versions of the models we have chosen are GPT-4o-0513 \cite{openai2024gpt4} for generation, and bge-large-en-v1.5 \cite{bge_embedding} for retrieval.

For efficiency and comparability, we conducted ablation experiments on the MIN DEV officially released by BIRD. The purpose of these ablation experiments is to evaluate the generalization ability of OpenSearch-SQL across various models, as well as to verify the effectiveness of each submodule in the method.
During the \textbf{Extraction}, we set the temperature from 0, in \textbf{Generation} and \textbf{Refine} stage, the temperature is set to 0.7. The number of few-shot examples is selected from $N=\{0,3,5,7,9\}$ , with a filtering index threshold value of $0.65$, and the maximum number of SQL votes is chosen from $K = \{1, 7, 15, 21\}$.

\subsection{Main Result}
To demonstrate the effectiveness of OpenSearch-SQL, our performance on the Spider and BIRD datasets has achieved state-of-the-art (SOTA) levels, highlighting the stability and superiority of our approach. In Table \ref{main result}, we compare the effectiveness of our method with the baseline methods in the BIRD dataset. Furthermore, Table \ref{main result2} presents the experimental results on the Spider dataset. Since the Spider leaderboard will no longer accept new submissions starting February 2024, the results are composed of current leaderboard data and those reported in relevant papers. Notably, the BIRD dataset underwent a cleansing of the devlopment set in July, which might lead to discrepancies in comparisons. To ensure comparability with previous evaluations, we have continued to use the dataset before July for our experiments.
 \paragraph{\textbf{BIRD Results}} We present the performance of Baseline and OpenSearch-SQL, v2 on the BIRD dataset in Table \ref{main result2}. The results in the table are derived entirely from the BIRD leaderboard. At the time of submission, our method achieved 69.3\% EX on the BIRD development set and 72.28\% EX and 69.36\% R-VES on the BIRD holdout test set, all ranking first. It is noteworthy that our method did not involve any fine-tuning to customize the LLM. Meanwhile, we demonstrated the performance of OpenSearch-SQL without applying Self-Consistency \& Vote. It can be seen that even using a single generation SQL, OpenSearch-SQL achieved an EX of 67.8, still maintains the top performance on the development set. 

\begin{figure}
     \centering
     \includegraphics[width=1.0\linewidth]{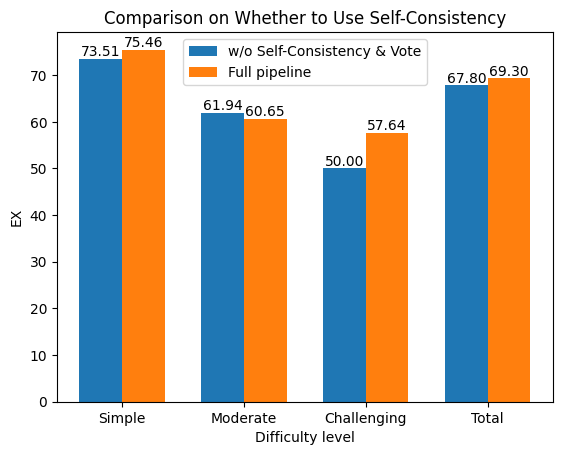}
     \caption{Comparison of the impact of using Consistency \& Vote across different difficulty levels.}
     \label{difficulty}
 \end{figure}

In Figure \ref{difficulty}, we compare the performance of our method across various difficulty levels. It is evident that Self-Consistency \& Vote shows the most significant improvement for difficult problems, achieving an absolute difference of 7.64\%. However, there is no notable difference for easy and medium problems. This indicates that as problem difficulty increases, the susceptibility of large models to hallucinations also rises.

\begin{table}[h]
\Large
\centering
\renewcommand\arraystretch{1.5}
\resizebox{\linewidth}{!}{
\begin{tabular}{c|cc|cc}
\hline
Method & \multicolumn{2}{c|}{EX} & \multicolumn{1}{c}{R-VES} \\
&   dev& test &  test \\ \hline \hline
GPT-4   &46.35& 54.89 & 51.57 \\
DIN-SQL + GPT-4  & 50.72 & 55.90 &  53.07\\ 
DAIL-SQL + GPT-4  & 54.76 & 57.41 & 54.02\\
MAC-SQL + GPT-4  & 57.56 & 59.59&57.60\\
MCS-SQL + GPT-4  & 63.36 &65.45 & 61.23\\
CHESS & 65.00 &66.69& 62.77\\
Distillery + GPT-4o(ft)  & 67.21 &71.83 & 67.41\\\hline
 OpenSearch-SQL + GPT-4& 66.62 & - &-\\
 OpenSearch-SQL + GPT-4o & 67.80 & - &-\\
 {\normalsize w/o Self-Consistency \& Vote}&  &  &\\
 \textbf{OpenSearch-SQL + GPT-4o}  &\textbf{69.30} & \textbf{72.28}&  \textbf{69.36}\\\hline
\end{tabular}
}
\caption{Execution accuracies (EX) and Reward-based Valid Efficiency Scores (R-VES) on the BIRD dev and test sets. The result is taken from the leaderboard.}\label{main result2}
\end{table}

\paragraph{\textbf{Spider Results}} To demonstrate the generalizability of OpenSearch-SQL, we also evaluated its performance on the Spider test set. We used the default configuration of our method, making no other changes, except for necessary adjustments to accommodate the data set format. Since the Spider leaderboard is no longer updated and the test set is publicly available, we extracted results from the baseline method's paper and leaderboard, with results extracted from the paper marked with a $*$. Table \ref{main result} shows the performance of the methods, where our method achieved an execution accuracy of 87.1\%.
\begin{table}[th]
\centering
\tiny
\renewcommand\arraystretch{1.2}
\resizebox{\linewidth}{!}{
\begin{tabular}{c|c}
\hline
Method & \multicolumn{1}{c}{EX}  \\
 \hline \hline
GPT-4    & 83.9  \\
C3+ChatGPT & 82.3 \\ 
 DIN-SQL  & 85.3 \\ 
 DAIL-SQL+GPT-4   & 86.6\\
 MAC-SQL+GPT-4  &82.8*\\
 MCS-SQL+GPT-4  &89.6* \\
 CHESS & 87.2*\\
 Distillery + GPT-4o(ft)  & -\\\hline
 OpenSearch-SQL+GPT-4  & 86.8\\
 \textbf{OpenSearch-SQL+GPT-4o}  & 87.1\\\hline
\end{tabular}
}
\caption{Execution accuracy (EX) for the Spider test sets. * denotes that the result is taken from their paper rather than from the leaderboard.}\label{main result}
\end{table}



\begin{table*}[h]
\centering
\renewcommand\arraystretch{1.5} 
\resizebox{\linewidth}{!}{
\begin{tabular}{c|cc|ccc|ccc}
\hline
Pipeline Setup & $EX_G$  & $\Delta EX_G$ & $EX_R$ & $\Delta EX_R$ &$EX_R-EX_G$& $EX$ & $\Delta EX$ & $EX-EX_R$ \\ \hline \hline
Full pipeline & 65.8 & - & 68.2 & -&$+2.4$ & 70.6 & -&$+2.4$ \\ \hline
w/o \textbf{Extraction} & 61.6 & $-4.2$ & 66.2 & $-2.0$ & $+4.6$ & 67.4 & $-3.2$&$+1.4$ \\ 
w/o Values Retrieval & 64.4 & $-1.4$ & 66.6 & $-1.6$& $+2.2$ & 69.2 & $-1.4$ &$+2.6$\\
w/o column filtering & 63.2 & $-2.6$ & 65.0 & $-3.2$ &$+1.8$& 68.6 & $-2.0 $&$+3.6$\\\hline 
w/o Info Alignment & 62.8 & $-3.0$ & 67.6 & $-0.6$&$+4.8$ & 68.6 & $-2.0 $&$+1.0 $\\ \hline
w/o \textbf{Generation} & - & - & - & - & - & - &-&-\\ 
w/o Few-shot & 60.4 & $-5.4$ & 63.0 & $-5.3$ &$+2.6$& 66.0 & $-4.6$&$+3.0$ \\
w/o CoT & 63.0 & $-2.8$ & 66.2 & $-2.0$ &$+3.2$& 69.2 & $-1.4$&$+2.6$ \\\hline
w/o Alignments & 65.8 & - & 67.0 &$-1.2$&$+1.2$  & 69.6 & $-1.0$ &$+2.6$\\ \hline
w/o \textbf{Refinement} & 65.8 & - & 67.0 &$-1.2$&$+1.2$ &67.0 &- &- \\ 
w/o Correction & 65.8 & - & 67.0 &$-1.2$ &$+1.0$ & 69.8 & $-0.8$&$+2.8$ \\ 
w/o Few-Shot & 65.8 & - & 67.6 & $-0.6$&$+1.6$ & 69.4 & $-1.2$&$+1.8$ \\ 
w/o Self-Consistency \& Vote & 65.8 & - & 68.2 & - &-& 68.2 & - &-\\ \hline \hline
\end{tabular}}
\caption{The execution accuracy (EX) of the pipeline by removing each component on the Mini-Dev set. $EX_G$ represents the accuracy of a single SQL obtained from the \textbf{Generation}, $EX_R$ indicates the performance of a single SQL before Self-Consistency \& voting, and EX refers to the results after all processes are completed.}
\label{component}
\end{table*}                                                                                
\subsection{Modular Ablation}

Table \ref{component} provides a detailed overview of the Execution Accuracy (EX) when each module is removed from the method. We specifically examine the roles of the \textbf{Extraction}, \textbf{Generation}, \textbf{Refinement}, and \textbf{Alignment} modules. Furthermore, we investigate the impact of key submodules such as Retrieve, Few-shot, Chain of Thought (CoT), and Self-consistency \& Vote. Notably, Self-consistency \& Vote functions as the final step in the selection among multiple SQLs. To more clearly illustrate the effect of each module on the optimization of individual SQL, we present the EX results of this portion separately.

To thoroughly demonstrate the roles of each module in the method, we first recorded the Execution Accuracy (EX) of the SQLs ultimately generated by OpenSearch-SQL. Given that the Consistency \& Vote mechanism selects from multiple SQLs, it may influence the evaluation of the impact of each module on an individual SQL. Therefore, we also recorded the accuracy of the execution of the SQLs obtained at the generation stage ($EX_G$) and prior to the Consistency \& Vote stage ($EX_R$). These two metrics directly capture the effect of each module on individual SQLs, providing a more precise assessment of each module's influence.

As shown in the table \ref{component}, from left to right, we present $EX_G$, $EX_R$, $EX$. From top to bottom, the roles of the Extraction, Generation, Refinement, and Alignment modules are recorded in sequence.
From the table, we can observe:
\begin{itemize}
    \item The EX of SQLs increases monotonically as the workflow progresses, indicating that different modules have distinct positive impacts on SQL optimization, confirming the necessity of each module.
    \item Without Extraction, the complete database schema is used directly, bypassing values retrieval and column filtering. It can be observed that the combined improvement from values retrieval and column filtering for a single SQL is approximately equal to the improvement provided by Extraction, suggesting minimal overlap in their roles.
    \item Few-shot demonstrates a significant effect during the generation phase and further enhances the final SQL performance during the Refinement phase, highlighting the importance of a dynamic Few-shot approach. We speculate that few-shot can greatly enhance the model's potential, while also improving the stability of the generated results. Further analysis of few-shot will be conducted in the experiments \ref{fewshot}.
    \item During the \textbf{Generation}, CoT and Few-shot significantly enhance individual SQL results, indicating a positive correlation between the amount of information in the prompt and the effectiveness of LLM-generated SQL. However, after the \textbf{Refinement}
    and Self-Consistency \& Vote stages, the gains from CoT are notably reduced. We believe this suggests that CoT does not significantly enhance the model's foundational generation capabilities but instead improves the stability of the results. Additionally, it is observed that \textbf{Refinement} provides $+3.2\%$ to results generated without CoT, highlighting its crucial role in increasing result stability.
    \item The Alignment module has played a positive role at various stages, significantly enhancing the performance of individual SQL, thereby demonstrating its unique optimization potential. Specifically, Alignments applied after the generation phase can assist the Correction module in accurately fixing erroneous SQL, rather than merely making them executable.
    \item It is evident that Self-Consistency \& Vote consistently raise the performance ceiling of LLMs. Based on our observations of the generated SQL, this mechanism primarily helps avoid errors caused by low-probability randomness.
\end{itemize}


\begin{table*}[h]
\large
\centering
\renewcommand\arraystretch{1.5}
\resizebox{\linewidth}{!}{
\begin{tabular}{c|cccccc}
\hline
 Method& $EX_{G}$  &  $\Delta EX_{G}$  & $EX_{R}$ &  $\Delta EX_{R}$ & EX & $\Delta EX$ \\ \hline \hline
Query-CoT-SQL pair Few-shot  & 65.8& -& 68.2& -& 70.6& -  \\\hline
w/o Few-shot of Generation & 59.6&$-6.2$ & 63.0&$-5.2$ & 66.0&$-4.6$   \\ 
w Query-SQL pair Few-shot of Generation & 63.0&$-2.8$& 66.2&$-2.0$ & 69.2&$-1.4$ \\ 
 w/o Few-shot of Refinement&  65.8&-&  67.6& $ -0.6$&69.4  &$-1.2$\\ 
 w/o Few-shot of Generation \& Refinement &59.6  &$-6.2$& 62.8 &$-5.4$& 66.0& $-4.6$ \\ 
\hline
\end{tabular}}
\caption{Few-shot Performance Comparison Results.}\label{fewshotexp}
\end{table*}

\subsection{Few-shot}\label{fewshot}

In Table \ref{fewshotexp}, we explore the impact of different dynamic Few-shot strategies on Text-to-SQL. First, during the generation phase, we compare the Few-shot effects of the classic Query-SQL pair with our Query-COT-SQL pair, and we evaluate the scenario without Few-shot. We also investigate the influence of Few-shot in other stages. From the table, we can summarize the following points:
\begin{itemize}
    \item The Few-shot strategy significantly raises the upper limit of SQL accuracy generated by LLMs. Whether in the form of a Query-SQL Pair or a Query-CoT-SQL Pair, compared to without few-shot, these formats significantly enhance the performance of LLMs at each stage. However, the improvement from few-shot in the \textbf{Refinement} stage is relatively smaller.
    \item During the \textbf{Generation}, for a single SQL query, the Few-shot strategy provides the greatest performance improvement, with the Query-CoT-SQL pair form achieving a 3.0\% absolute EX increase compared to the Query-SQL pair form, and a 6.4\% increase in EX compared to the scenario without Few-shot.
    \item In \textbf{Refine} and under the influence of \textbf{self-consistency \& vote}, the gap between different forms of Few-shot gradually narrows, indicating that there is some redundancy in the SQL optimization content in different stages.
    \item Although the Few-shot strategy has limited optimization effects on individual SQL queries during the \textbf{Refine}, it can improve the final EX of consistency and voting, suggesting that the Few-shot-driven SQL correction results exhibit higher consistency with one another.
\end{itemize}

\subsection{Self-Consistency \& Vote}
\begin{figure}[h]
    \centering
    \includegraphics[width=1.0\linewidth]{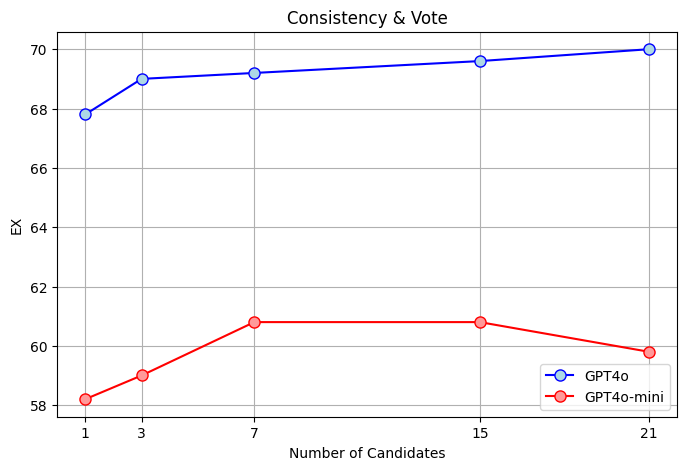}
    \caption{Comparison of the impact of different numbers of candidates on model performance}
    \label{fig:enter-label}
\end{figure}

In our result, we used a single prompt to generate 21 results as answers. To investigate the optimal number of SQL outputs generated by beam search for self-consistency with a single prompt, we designed the experiment shown in the figure \ref{fig:enter-label}. 

We conducted experiment with GPT-4o and GPT-4o-mini to compare the performance of LLMs with different parameter levels, and the number of candidates is taken from $N=\{1, 3, 7,15, 21\}$. As observed in the figure, the EX value for GPT-4o consistently increases with the number of candidates. In contrast, GPT-4o-mini achieves optimal results with 7 and 15 candidates. This indicates that for models with sufficiently large parameters, the number of generated answers can be safely increased to achieve self-consistency \& vote. However, for smaller models, the number of outputs should be controlled to prevent negative impacts on self-consistency due to an excessive number of answers.

\subsection{Execution Cost}

In Table \ref{cost}, we provide a simple estimate of the time and resource consumption for using OpenSearch-SQL with GPT-4o. Generally, the execution time for the entire process primarily depends on large language model (LLM) calls, retrieval time, and SQL execution time. In practice, using HNSW \cite{hnsw} can significantly reduce retrieval latency, thus shifting the main time consumption to LLM responses and SQL execution.

Specifically, the tokens for extraction and generation are mainly derived from the database schema and a few examples. The generation phase takes longer since it uses BeamSearch, which generates more tokens than the extraction phase. Apart from the SELECT alignment, which is triggered every time, other alignment steps are only activated when there are issues with the SQL, so they are not time-consuming in most cases. As for the refinement phase, the time is mainly due to SQL execution and LLM corrections, while Self-consistency \& Vote contribute minimally to the time consumption.

\begin{table}[th]
\centering
\tiny
\renewcommand\arraystretch{1.2}
\resizebox{\linewidth}{!}{
\begin{tabular}{c|cc}
\hline
 Modular& Time(s) & Token \\ \hline
\textbf{Extraction} &4-9 &5000-10000\\
Entity \& Column & 4-6& 5000-10000\\ 
Retrieval & 0-1& -\\\hline
\textbf{Generation}& 5-15 &4000-8000\\ \hline
\textbf{Refinement}& 0-25&0-5000\\
Correction& 0-25&0-5000\\
Self-consistency \& Vote & <0.01s &-\\ \hline
\textbf{Alignments}&0-15 & 500-2000\\
SELECT Alignment& 1-3& 500-600\\
Agent Alignment &0-7 & 100-500\\
Style Alignment &0-5& 100-500\\
Function Alignment &0-4 &100-500\\ \hline
Pipeline&7-60s & 9000-25000\\ \hline

\end{tabular}
}
\caption{Exection Cost Performance}\label{cost}
\end{table}

\subsection{CoT}

In this section, we provide a brief analysis of the CoT method we designed and compare the performance of structured CoT designed for OpenSearch-SQL with the unstructured CoT approach( i.e., "let's think step by step") as well as scenarios without CoT. To isolate the impact of CoT, we eliminate few-shot learning during the \textbf{Generation} phase and focus solely on the execution accuracy when generating a single SQL( $EX_G$) and multiple SQLs with Self-consistency \& Vote( $EX_V$). This allows for a straightforward assessment of CoT's direct impact on SQL generation.

As shown in Table \ref{CoT},t is apparent that utilizing CoT leads to a significant improvement in SQL generation compared to not using CoT. Moreover, structured CoT demonstrates more substantial gains than unstructured CoT. Furthermore, examining the results of $EX_{V}-EX_G$, we observe that the implementation of \textbf{Self-consistency \& Vote} provides even greater improvements than simply using CoT alone, with structured CoT achieving a higher relative enhancement.

In summary, structured CoT achieved the optimal results. When sampling multiple answers for voting, structured CoT exhibits a more significant relative improvement compared to other methods.

\begin{table}[th]
\centering
\tiny
\renewcommand\arraystretch{1.2}
\resizebox{\linewidth}{!}{
\begin{tabular}{c|ccc}
\hline
 Modular& $EX_G$ & $EX_{V}$ &$EX_{V}-EX_G$\\ \hline
 w/o CoT& 57.6&59.2 & 1.6\\ 
 Unstructured CoT &58.2 &63.0&4.8 \\
 Structured CoT& 58.8 &65.0 &6.2\\ \hline
\end{tabular}
}
\caption{CoT Performance Comparison Results}\label{CoT}
\end{table}

\section{Related Work}
For a long time, computer professionals have been hoping to convert natural language queries into specific database queries according to certain scenarios, in order to dramatically reduce the learning and usage costs for individuals, bridge the gap between laypersons and expert users, and enhance the efficiency of using, analyzing, and querying databases. The Text-to-SQL task is one of the most representative examples of these, and success in the Text-to-SQL task would imply that similar methods could assist in various SQL-like tasks. However, this task continues to be challenging, particularly due to the strong correlation between the SQL structure and the database structure, as well as the limitations imposed by the expression format requirements\cite{katsogiannis2023survey}.

\subsection{Classic Method}
Classical methods often do not directly generate SQL but assemble it through various means. These methods handle components like SELECT, WHERE, etc., separately to construct SQL clauses. An example of such a method is SQLNET\cite{sqlnet}, which employs a sketch-based approach where the sketch contains a dependency graph, allowing one prediction to take into account only the previous predictions it depends on. RYANSQL\cite{ryansql} and SyntaxSQLNet\cite{syntaxsqlnet} have tried to generalize sketch-based decoding, attempting not only to fill in the gaps of a sketch but also to generate the appropriate sketch for a given text. Additionally, there are methods that utilize graph representation\cite{graph_1,graph2} and employ intermediate languages\cite{irnet,smbop} to enhance the capabilities of LLMs. These methods may be constrained by the inherent capabilities of the models, which may lead to inferior performance. For the Text-to-SQL task, their theoretical value outweighs their practical value.

\subsection{Large Language Models}
Large Language Models (LLMs) represent a significant breakthrough in the field of artificial intelligence in recent years. They are based on deep learning technology, typically built on the Transformer architecture. These models consist of large-scale neural networks capable of training on massive amounts of text data, thereby endowing them with the powerful ability to generate and understand natural language text.

Throughout the development of LLMs, several notable models have emerged, the most famous being GPT (Generative Pre-trained Transformer) \cite{gpt} and BERT (Bidirectional Encoder Representations from Transformers) \cite{bert}. GPT excels in natural language generation, while BERT is exceptional in understanding natural language. These models are widely applied across various fields, such as text generation, information extraction, and machine translation, effectively advancing natural language processing.

Recently, more advanced models have emerged, such as GPT-4 \cite{openai2024gpt4}, LLAMA \cite{llama3}, Qwenmax \cite{bai2023qwen}, and DeepSeekV2.5 \cite{liu2024deepseek}, which features a mixture of experts (MoE) architecture . These emerging models can be regarded as significant milestones in the fields of natural language processing and machine learning due to their outstanding performance in a variety of downstream tasks. Thanks to the excellent capabilities of these models, natural language processing now plays an increasingly important role in addressing many complex and meaningful issues. As technology continues to advance, large language models are expected to have a profound impact on scientific research, commercial applications, and daily life worldwide.

\subsection{LLM-based Agents and RAG}
With the advent of the era of large models, the enhanced capabilities of large models have simplified many previously difficult tasks. The LLM-based agent is an approach that leverages the power of large models to carry out a multitude of complex tasks\cite{llmagentsurvey}, usually performing tasks by assuming specific roles, such as programmers, educators, and domain experts\cite{agents1}. Retrieval-Augmented Generation (RAG) is a type of agent that assists in answering questions by retrieving highly relevant documents\cite{rag}. Furthermore, multiple RAG techniques, including Modular RAG\cite{modrag1,modrag2} and Advanced RAG\cite{advance_rag1,advrag2}, assist LLMs in providing more accurate and richly formatted answers by employing techniques such as re-ranking and rewriting.

\subsection{LLM for Text-to-SQL}
Recent research shows that the most effective Text-to-SQL methods primarily rely on multi-Agent algorithms based on LLMs. These methods aim to provide the most helpful information to the LLM through Agents to assist in predicting SQL. In this process, the classic Text-to-SQL task benchmarks have gradually evolved and become more complex, with commonly used Text-to-SQL evaluation tasks progressively shifting from WiKiSQL\cite{wikisql} to Spider\cite{spider} and BIRD\cite{Bird}.
In these tasks, researchers\cite{tai2023exploring} enhance the LLM's Text-to-SQL abilities through chain-of-thought reasoning. C3\cite{c3} has developed a systematic zero-shot Text-to-SQL method that includes several key components. DIN-SQL\cite{dinsql} addresses the Text-to-SQL task by distinguishing question difficulty and adjusting specific prompts accordingly. DAIL-SQL\cite{DAILsQL} assists the model in generating SQL by retrieving structurally similar questions or SQL as few-shot examples, while MAC-SQL improves capabilities through task decomposition and recombination. Meanwhile, some studies have opted for a fine-tuning + Agent approach to accomplish the Text-to-SQL task, such as Dubo-SQL\cite{dubo}, SFT CodeS\cite{codes}, DTS-SQL\cite{dtssql}. Notably powerful methods currently include MCS-SQL\cite{mcssql}, which enhances the stability of Text-to-SQL tasks by using multiple sets of prompts for voting. CHESS\cite{chess}, which simplifies the generation difficulty through an excellent column filtering mechanism, and Distillery\cite{Distilgpt4}, which optimizes the task by fine-tuning GPT-4O.

Overall, as the capabilities of large models continue to improve, the framework for the Text-to-SQL task has gradually converged and standardized into the form of LLMs + Agents. Meanwhile, the capabilities of Text-to-SQL methods are also constantly enhancing, becoming tools that can truly help simplify people's work in everyday applications.

\section{Conclusion}
\label{sec:conclusion}
In this paper, we introduce OpenSearch-SQL, a method that enhances the performance of the Text-to-SQL task using dynamic Few-shot and consistency alignment mechanisms. To improve the model's handling of prompts, we extend the original examples with a LLM and supplement them with CoT information, forming a Query-CoT-SQL Few-shot configuration. To the best of our knowledge, this is the first exploration of using LLMs to extend Few-shot CoT content in the Text-to-SQL task. Additionally, we developed a novel method based on a consistency alignment mechanism to mitigate hallucinations, enhancing the quality of agent outputs by reintegrating their inputs and outputs. Our approach does not rely on any SFT tasks and is entirely based on directly available LLMs and retrieval models, representing a pure architecture upgrade for Text-to-SQL. As a result, at the time of submission, we achieved the top position in all three metrics on the BIRD leaderboard, which demonstrates the significant advantages of our approach.

Our source code will be made available soon, we hope that this Few-shot construction method and consistency alignment-based workflow will offer new perspectives and positive impacts for Text-to-SQL and other multi-agent collaborative tasks.

\bibliographystyle{ACM-Reference-Format}
\bibliography{main}


\begin{thebibliography}{48}


\ifx \showCODEN    \undefined \def \showCODEN     #1{\unskip}     \fi
\ifx \showDOI      \undefined \def \showDOI       #1{#1}\fi
\ifx \showISBNx    \undefined \def \showISBNx     #1{\unskip}     \fi
\ifx \showISBNxiii \undefined \def \showISBNxiii  #1{\unskip}     \fi
\ifx \showISSN     \undefined \def \showISSN      #1{\unskip}     \fi
\ifx \showLCCN     \undefined \def \showLCCN      #1{\unskip}     \fi
\ifx \shownote     \undefined \def \shownote      #1{#1}          \fi
\ifx \showarticletitle \undefined \def \showarticletitle #1{#1}   \fi
\ifx \showURL      \undefined \def \showURL       {\relax}        \fi
\providecommand\bibfield[2]{#2}
\providecommand\bibinfo[2]{#2}
\providecommand\natexlab[1]{#1}
\providecommand\showeprint[2][]{arXiv:#2}

\bibitem[Bai et~al\mbox{.}(2023)]%
        {bai2023qwen}
\bibfield{author}{\bibinfo{person}{Jinze Bai}, \bibinfo{person}{Shuai Bai}, \bibinfo{person}{Yunfei Chu}, \bibinfo{person}{Zeyu Cui}, \bibinfo{person}{Kai Dang}, \bibinfo{person}{Xiaodong Deng}, \bibinfo{person}{Yang Fan}, \bibinfo{person}{Wenbin Ge}, \bibinfo{person}{Yu Han}, \bibinfo{person}{Fei Huang}, {et~al\mbox{.}}} \bibinfo{year}{2023}\natexlab{}.
\newblock \showarticletitle{Qwen technical report}.
\newblock \bibinfo{journal}{\emph{arXiv preprint arXiv:2309.16609}} (\bibinfo{year}{2023}).
\newblock


\bibitem[Bogin et~al\mbox{.}(2019)]%
        {graph_1}
\bibfield{author}{\bibinfo{person}{Ben Bogin}, \bibinfo{person}{Matt Gardner}, {and} \bibinfo{person}{Jonathan Berant}.} \bibinfo{year}{2019}\natexlab{}.
\newblock \showarticletitle{Global reasoning over database structures for text-to-sql parsing}.
\newblock \bibinfo{journal}{\emph{arXiv preprint arXiv:1908.11214}} (\bibinfo{year}{2019}).
\newblock


\bibitem[Brunner and Stockinger(2021)]%
        {graph2}
\bibfield{author}{\bibinfo{person}{Ursin Brunner} {and} \bibinfo{person}{Kurt Stockinger}.} \bibinfo{year}{2021}\natexlab{}.
\newblock \showarticletitle{Valuenet: A natural language-to-sql system that learns from database information}. In \bibinfo{booktitle}{\emph{2021 IEEE 37th International Conference on Data Engineering (ICDE)}}. IEEE, \bibinfo{pages}{2177--2182}.
\newblock


\bibitem[Choi et~al\mbox{.}(2021)]%
        {ryansql}
\bibfield{author}{\bibinfo{person}{DongHyun Choi}, \bibinfo{person}{Myeong~Cheol Shin}, \bibinfo{person}{EungGyun Kim}, {and} \bibinfo{person}{Dong~Ryeol Shin}.} \bibinfo{year}{2021}\natexlab{}.
\newblock \showarticletitle{Ryansql: Recursively applying sketch-based slot fillings for complex text-to-sql in cross-domain databases}.
\newblock \bibinfo{journal}{\emph{Computational Linguistics}} \bibinfo{volume}{47}, \bibinfo{number}{2} (\bibinfo{year}{2021}), \bibinfo{pages}{309--332}.
\newblock


\bibitem[Devlin et~al\mbox{.}(2019)]%
        {bert}
\bibfield{author}{\bibinfo{person}{Jacob Devlin}, \bibinfo{person}{Ming-Wei Chang}, \bibinfo{person}{Kenton Lee}, {and} \bibinfo{person}{Kristina Toutanova}.} \bibinfo{year}{2019}\natexlab{}.
\newblock \bibinfo{title}{BERT: Pre-training of Deep Bidirectional Transformers for Language Understanding}.
\newblock
\newblock
\showeprint[arxiv]{1810.04805}~[cs.CL]
\urldef\tempurl%
\url{https://arxiv.org/abs/1810.04805}
\showURL{%
\tempurl}


\bibitem[Dong et~al\mbox{.}(2023)]%
        {c3}
\bibfield{author}{\bibinfo{person}{Xuemei Dong}, \bibinfo{person}{Chao Zhang}, \bibinfo{person}{Yuhang Ge}, \bibinfo{person}{Yuren Mao}, \bibinfo{person}{Yunjun Gao}, \bibinfo{person}{Jinshu Lin}, \bibinfo{person}{Dongfang Lou}, {et~al\mbox{.}}} \bibinfo{year}{2023}\natexlab{}.
\newblock \showarticletitle{C3: Zero-shot text-to-sql with chatgpt}.
\newblock \bibinfo{journal}{\emph{arXiv preprint arXiv:2307.07306}} (\bibinfo{year}{2023}).
\newblock


\bibitem[Gao et~al\mbox{.}(2023)]%
        {DAILsQL}
\bibfield{author}{\bibinfo{person}{Dawei Gao}, \bibinfo{person}{Haibin Wang}, \bibinfo{person}{Yaliang Li}, \bibinfo{person}{Xiuyu Sun}, \bibinfo{person}{Yichen Qian}, \bibinfo{person}{Bolin Ding}, {and} \bibinfo{person}{Jingren Zhou}.} \bibinfo{year}{2023}\natexlab{}.
\newblock \showarticletitle{Text-to-sql empowered by large language models: A benchmark evaluation}.
\newblock \bibinfo{journal}{\emph{arXiv preprint arXiv:2308.15363}} (\bibinfo{year}{2023}).
\newblock


\bibitem[Gao et~al\mbox{.}(2022)]%
        {advrag2}
\bibfield{author}{\bibinfo{person}{Luyu Gao}, \bibinfo{person}{Xueguang Ma}, \bibinfo{person}{Jimmy Lin}, {and} \bibinfo{person}{Jamie Callan}.} \bibinfo{year}{2022}\natexlab{}.
\newblock \showarticletitle{Precise zero-shot dense retrieval without relevance labels}.
\newblock \bibinfo{journal}{\emph{arXiv preprint arXiv:2212.10496}} (\bibinfo{year}{2022}).
\newblock


\bibitem[Gao et~al\mbox{.}(2024)]%
        {rag}
\bibfield{author}{\bibinfo{person}{Yunfan Gao}, \bibinfo{person}{Yun Xiong}, \bibinfo{person}{Xinyu Gao}, \bibinfo{person}{Kangxiang Jia}, \bibinfo{person}{Jinliu Pan}, \bibinfo{person}{Yuxi Bi}, \bibinfo{person}{Yi Dai}, \bibinfo{person}{Jiawei Sun}, \bibinfo{person}{Meng Wang}, {and} \bibinfo{person}{Haofen Wang}.} \bibinfo{year}{2024}\natexlab{}.
\newblock \bibinfo{title}{Retrieval-Augmented Generation for Large Language Models: A Survey}.
\newblock
\newblock
\showeprint[arxiv]{2312.10997}~[cs.CL]


\bibitem[Guo et~al\mbox{.}(2023)]%
        {mqs}
\bibfield{author}{\bibinfo{person}{Chunxi Guo}, \bibinfo{person}{Zhiliang Tian}, \bibinfo{person}{Jintao Tang}, \bibinfo{person}{Pancheng Wang}, \bibinfo{person}{Zhihua Wen}, \bibinfo{person}{Kang Yang}, {and} \bibinfo{person}{Ting Wang}.} \bibinfo{year}{2023}\natexlab{}.
\newblock \bibinfo{title}{Prompting GPT-3.5 for Text-to-SQL with De-semanticization and Skeleton Retrieval}.
\newblock
\newblock
\showeprint[arxiv]{2304.13301}~[cs.CL]
\urldef\tempurl%
\url{https://arxiv.org/abs/2304.13301}
\showURL{%
\tempurl}


\bibitem[Guo et~al\mbox{.}(2019a)]%
        {nmt1}
\bibfield{author}{\bibinfo{person}{Jiaqi Guo}, \bibinfo{person}{Zecheng Zhan}, \bibinfo{person}{Yan Gao}, \bibinfo{person}{Yan Xiao}, \bibinfo{person}{Jian-Guang Lou}, \bibinfo{person}{Ting Liu}, {and} \bibinfo{person}{Dongmei Zhang}.} \bibinfo{year}{2019}\natexlab{a}.
\newblock \showarticletitle{Towards complex text-to-sql in cross-domain database with intermediate representation}.
\newblock \bibinfo{journal}{\emph{arXiv preprint arXiv:1905.08205}} (\bibinfo{year}{2019}).
\newblock


\bibitem[Guo et~al\mbox{.}(2019b)]%
        {irnet}
\bibfield{author}{\bibinfo{person}{Jiaqi Guo}, \bibinfo{person}{Zecheng Zhan}, \bibinfo{person}{Yan Gao}, \bibinfo{person}{Yan Xiao}, \bibinfo{person}{Jian-Guang Lou}, \bibinfo{person}{Ting Liu}, {and} \bibinfo{person}{Dongmei Zhang}.} \bibinfo{year}{2019}\natexlab{b}.
\newblock \showarticletitle{Towards complex text-to-sql in cross-domain database with intermediate representation}.
\newblock \bibinfo{journal}{\emph{arXiv preprint arXiv:1905.08205}} (\bibinfo{year}{2019}).
\newblock


\bibitem[He et~al\mbox{.}(2015)]%
        {resnet}
\bibfield{author}{\bibinfo{person}{Kaiming He}, \bibinfo{person}{Xiangyu Zhang}, \bibinfo{person}{Shaoqing Ren}, {and} \bibinfo{person}{Jian Sun}.} \bibinfo{year}{2015}\natexlab{}.
\newblock \bibinfo{title}{Deep Residual Learning for Image Recognition}.
\newblock
\newblock
\showeprint[arxiv]{1512.03385}~[cs.CV]
\urldef\tempurl%
\url{https://arxiv.org/abs/1512.03385}
\showURL{%
\tempurl}


\bibitem[Hristidis and Papakonstantinou(2002)]%
        {earlytxt2sql2}
\bibfield{author}{\bibinfo{person}{Vagelis Hristidis} {and} \bibinfo{person}{Yannis Papakonstantinou}.} \bibinfo{year}{2002}\natexlab{}.
\newblock \showarticletitle{Discover: Keyword search in relational databases}. In \bibinfo{booktitle}{\emph{VLDB'02: Proceedings of the 28th International Conference on Very Large Databases}}. Elsevier, \bibinfo{pages}{670--681}.
\newblock


\bibitem[Hristidis et~al\mbox{.}(2003)]%
        {earlytxt2sql1}
\bibfield{author}{\bibinfo{person}{Vagelis Hristidis}, \bibinfo{person}{Yannis Papakonstantinou}, {and} \bibinfo{person}{Luis Gravano}.} \bibinfo{year}{2003}\natexlab{}.
\newblock \showarticletitle{Efficient IR-style keyword search over relational databases}. In \bibinfo{booktitle}{\emph{Proceedings 2003 VLDB Conference}}. Elsevier, \bibinfo{pages}{850--861}.
\newblock


\bibitem[Iyer et~al\mbox{.}(2017)]%
        {parse_survey1}
\bibfield{author}{\bibinfo{person}{Srinivasan Iyer}, \bibinfo{person}{Ioannis Konstas}, \bibinfo{person}{Alvin Cheung}, \bibinfo{person}{Jayant Krishnamurthy}, {and} \bibinfo{person}{Luke Zettlemoyer}.} \bibinfo{year}{2017}\natexlab{}.
\newblock \showarticletitle{Learning a neural semantic parser from user feedback}.
\newblock \bibinfo{journal}{\emph{arXiv preprint arXiv:1704.08760}} (\bibinfo{year}{2017}).
\newblock


\bibitem[Katsogiannis-Meimarakis and Koutrika(2023)]%
        {katsogiannis2023survey}
\bibfield{author}{\bibinfo{person}{George Katsogiannis-Meimarakis} {and} \bibinfo{person}{Georgia Koutrika}.} \bibinfo{year}{2023}\natexlab{}.
\newblock \showarticletitle{A survey on deep learning approaches for text-to-SQL}.
\newblock \bibinfo{journal}{\emph{The VLDB Journal}} \bibinfo{volume}{32}, \bibinfo{number}{4} (\bibinfo{year}{2023}), \bibinfo{pages}{905--936}.
\newblock


\bibitem[Lee et~al\mbox{.}(2024)]%
        {mcssql}
\bibfield{author}{\bibinfo{person}{Dongjun Lee}, \bibinfo{person}{Choongwon Park}, \bibinfo{person}{Jaehyuk Kim}, {and} \bibinfo{person}{Heesoo Park}.} \bibinfo{year}{2024}\natexlab{}.
\newblock \showarticletitle{MCS-SQL: Leveraging Multiple Prompts and Multiple-Choice Selection For Text-to-SQL Generation}.
\newblock \bibinfo{journal}{\emph{arXiv preprint arXiv:2405.07467}} (\bibinfo{year}{2024}).
\newblock


\bibitem[Li et~al\mbox{.}(2024b)]%
        {codes}
\bibfield{author}{\bibinfo{person}{Haoyang Li}, \bibinfo{person}{Jing Zhang}, \bibinfo{person}{Hanbing Liu}, \bibinfo{person}{Ju Fan}, \bibinfo{person}{Xiaokang Zhang}, \bibinfo{person}{Jun Zhu}, \bibinfo{person}{Renjie Wei}, \bibinfo{person}{Hongyan Pan}, \bibinfo{person}{Cuiping Li}, {and} \bibinfo{person}{Hong Chen}.} \bibinfo{year}{2024}\natexlab{b}.
\newblock \bibinfo{title}{CodeS: Towards Building Open-source Language Models for Text-to-SQL}.
\newblock
\newblock
\showeprint[arxiv]{2402.16347}~[cs.CL]


\bibitem[Li et~al\mbox{.}(2024a)]%
        {Bird}
\bibfield{author}{\bibinfo{person}{Jinyang Li}, \bibinfo{person}{Binyuan Hui}, \bibinfo{person}{Ge Qu}, \bibinfo{person}{Jiaxi Yang}, \bibinfo{person}{Binhua Li}, \bibinfo{person}{Bowen Li}, \bibinfo{person}{Bailin Wang}, \bibinfo{person}{Bowen Qin}, \bibinfo{person}{Ruiying Geng}, \bibinfo{person}{Nan Huo}, {et~al\mbox{.}}} \bibinfo{year}{2024}\natexlab{a}.
\newblock \showarticletitle{Can llm already serve as a database interface? a big bench for large-scale database grounded text-to-sqls}.
\newblock \bibinfo{journal}{\emph{Advances in Neural Information Processing Systems}}  \bibinfo{volume}{36} (\bibinfo{year}{2024}).
\newblock


\bibitem[Liu et~al\mbox{.}(2024)]%
        {liu2024deepseek}
\bibfield{author}{\bibinfo{person}{Aixin Liu}, \bibinfo{person}{Bei Feng}, \bibinfo{person}{Bin Wang}, \bibinfo{person}{Bingxuan Wang}, \bibinfo{person}{Bo Liu}, \bibinfo{person}{Chenggang Zhao}, \bibinfo{person}{Chengqi Dengr}, \bibinfo{person}{Chong Ruan}, \bibinfo{person}{Damai Dai}, \bibinfo{person}{Daya Guo}, {et~al\mbox{.}}} \bibinfo{year}{2024}\natexlab{}.
\newblock \showarticletitle{Deepseek-v2: A strong, economical, and efficient mixture-of-experts language model}.
\newblock \bibinfo{journal}{\emph{arXiv preprint arXiv:2405.04434}} (\bibinfo{year}{2024}).
\newblock


\bibitem[Ma et~al\mbox{.}(2023)]%
        {advance_rag1}
\bibfield{author}{\bibinfo{person}{Xinbei Ma}, \bibinfo{person}{Yeyun Gong}, \bibinfo{person}{Pengcheng He}, \bibinfo{person}{Hai Zhao}, {and} \bibinfo{person}{Nan Duan}.} \bibinfo{year}{2023}\natexlab{}.
\newblock \showarticletitle{Query rewriting for retrieval-augmented large language models}.
\newblock \bibinfo{journal}{\emph{arXiv preprint arXiv:2305.14283}} (\bibinfo{year}{2023}).
\newblock


\bibitem[Maamari et~al\mbox{.}(2024)]%
        {Distilgpt4}
\bibfield{author}{\bibinfo{person}{Karime Maamari}, \bibinfo{person}{Fadhil Abubaker}, \bibinfo{person}{Daniel Jaroslawicz}, {and} \bibinfo{person}{Amine Mhedhbi}.} \bibinfo{year}{2024}\natexlab{}.
\newblock \bibinfo{title}{The Death of Schema Linking? Text-to-SQL in the Age of Well-Reasoned Language Models}.
\newblock
\newblock
\showeprint[arxiv]{2408.07702}~[cs.CL]
\urldef\tempurl%
\url{https://arxiv.org/abs/2408.07702}
\showURL{%
\tempurl}


\bibitem[Malkov and Yashunin(2018)]%
        {hnsw}
\bibfield{author}{\bibinfo{person}{Yu.~A. Malkov} {and} \bibinfo{person}{D.~A. Yashunin}.} \bibinfo{year}{2018}\natexlab{}.
\newblock \bibinfo{title}{Efficient and robust approximate nearest neighbor search using Hierarchical Navigable Small World graphs}.
\newblock
\newblock
\showeprint[arxiv]{1603.09320}~[cs.DS]
\urldef\tempurl%
\url{https://arxiv.org/abs/1603.09320}
\showURL{%
\tempurl}


\bibitem[Meta(2024)]%
        {llama3}
\bibfield{author}{\bibinfo{person}{Meta}.} \bibinfo{year}{2024}\natexlab{}.
\newblock \bibinfo{title}{Introducing Meta Llama 3: The most capable openly available LLM to date}.
\newblock \bibinfo{howpublished}{\url{https://ai.meta.com/blog/meta-llama-3/}}.
\newblock
\newblock
\shownote{Accessed: 2023-05-20}.


\bibitem[OpenAI et~al\mbox{.}(2024)]%
        {openai2024gpt4}
\bibfield{author}{\bibinfo{person}{OpenAI}, \bibinfo{person}{Josh Achiam}, \bibinfo{person}{Steven Adler}, \bibinfo{person}{Sandhini Agarwal}, \bibinfo{person}{Lama Ahmad}, \bibinfo{person}{Ilge Akkaya}, \bibinfo{person}{Florencia~Leoni Aleman}, \bibinfo{person}{Diogo Almeida}, \bibinfo{person}{Janko Altenschmidt}, \bibinfo{person}{Sam Altman}, {et~al\mbox{.}}} \bibinfo{year}{2024}\natexlab{}.
\newblock \bibinfo{title}{GPT-4 Technical Report}.
\newblock
\newblock
\showeprint[arxiv]{2303.08774}~[cs.CL]
\urldef\tempurl%
\url{https://arxiv.org/abs/2303.08774}
\showURL{%
\tempurl}


\bibitem[Pourreza and Rafiei(2024a)]%
        {dinsql}
\bibfield{author}{\bibinfo{person}{Mohammadreza Pourreza} {and} \bibinfo{person}{Davood Rafiei}.} \bibinfo{year}{2024}\natexlab{a}.
\newblock \showarticletitle{Din-sql: Decomposed in-context learning of text-to-sql with self-correction}.
\newblock \bibinfo{journal}{\emph{Advances in Neural Information Processing Systems}}  \bibinfo{volume}{36} (\bibinfo{year}{2024}).
\newblock


\bibitem[Pourreza and Rafiei(2024b)]%
        {dtssql}
\bibfield{author}{\bibinfo{person}{Mohammadreza Pourreza} {and} \bibinfo{person}{Davood Rafiei}.} \bibinfo{year}{2024}\natexlab{b}.
\newblock \bibinfo{title}{DTS-SQL: Decomposed Text-to-SQL with Small Large Language Models}.
\newblock
\newblock
\showeprint[arxiv]{2402.01117}~[cs.CL]


\bibitem[Qian et~al\mbox{.}(2023)]%
        {agents1}
\bibfield{author}{\bibinfo{person}{Chen Qian}, \bibinfo{person}{Xin Cong}, \bibinfo{person}{Cheng Yang}, \bibinfo{person}{Weize Chen}, \bibinfo{person}{Yusheng Su}, \bibinfo{person}{Juyuan Xu}, \bibinfo{person}{Zhiyuan Liu}, {and} \bibinfo{person}{Maosong Sun}.} \bibinfo{year}{2023}\natexlab{}.
\newblock \showarticletitle{Communicative agents for software development}.
\newblock \bibinfo{journal}{\emph{arXiv preprint arXiv:2307.07924}} (\bibinfo{year}{2023}).
\newblock


\bibitem[Radford et~al\mbox{.}(2019)]%
        {gpt}
\bibfield{author}{\bibinfo{person}{Alec Radford}, \bibinfo{person}{Jeffrey Wu}, \bibinfo{person}{Rewon Child}, \bibinfo{person}{David Luan}, \bibinfo{person}{Dario Amodei}, \bibinfo{person}{Ilya Sutskever}, {et~al\mbox{.}}} \bibinfo{year}{2019}\natexlab{}.
\newblock \showarticletitle{Language models are unsupervised multitask learners}.
\newblock \bibinfo{journal}{\emph{OpenAI blog}} \bibinfo{volume}{1}, \bibinfo{number}{8} (\bibinfo{year}{2019}), \bibinfo{pages}{9}.
\newblock


\bibitem[Rubin and Berant(2020)]%
        {smbop}
\bibfield{author}{\bibinfo{person}{Ohad Rubin} {and} \bibinfo{person}{Jonathan Berant}.} \bibinfo{year}{2020}\natexlab{}.
\newblock \showarticletitle{SmBoP: Semi-autoregressive bottom-up semantic parsing}.
\newblock \bibinfo{journal}{\emph{arXiv preprint arXiv:2010.12412}} (\bibinfo{year}{2020}).
\newblock


\bibitem[Shao et~al\mbox{.}(2023)]%
        {modrag2}
\bibfield{author}{\bibinfo{person}{Zhihong Shao}, \bibinfo{person}{Yeyun Gong}, \bibinfo{person}{Yelong Shen}, \bibinfo{person}{Minlie Huang}, \bibinfo{person}{Nan Duan}, {and} \bibinfo{person}{Weizhu Chen}.} \bibinfo{year}{2023}\natexlab{}.
\newblock \showarticletitle{Enhancing retrieval-augmented large language models with iterative retrieval-generation synergy}.
\newblock \bibinfo{journal}{\emph{arXiv preprint arXiv:2305.15294}} (\bibinfo{year}{2023}).
\newblock


\bibitem[Tai et~al\mbox{.}(2023)]%
        {tai2023exploring}
\bibfield{author}{\bibinfo{person}{Chang-You Tai}, \bibinfo{person}{Ziru Chen}, \bibinfo{person}{Tianshu Zhang}, \bibinfo{person}{Xiang Deng}, {and} \bibinfo{person}{Huan Sun}.} \bibinfo{year}{2023}\natexlab{}.
\newblock \bibinfo{title}{Exploring Chain-of-Thought Style Prompting for Text-to-SQL}.
\newblock
\newblock
\showeprint[arxiv]{2305.14215}~[cs.CL]


\bibitem[Talaei et~al\mbox{.}(2024)]%
        {chess}
\bibfield{author}{\bibinfo{person}{Shayan Talaei}, \bibinfo{person}{Mohammadreza Pourreza}, \bibinfo{person}{Yu-Chen Chang}, \bibinfo{person}{Azalia Mirhoseini}, {and} \bibinfo{person}{Amin Saberi}.} \bibinfo{year}{2024}\natexlab{}.
\newblock \bibinfo{title}{CHESS: Contextual Harnessing for Efficient SQL Synthesis}.
\newblock
\newblock
\showeprint[arxiv]{2405.16755}~[cs.LG]
\urldef\tempurl%
\url{https://arxiv.org/abs/2405.16755}
\showURL{%
\tempurl}


\bibitem[Thorpe et~al\mbox{.}(2024)]%
        {dubo}
\bibfield{author}{\bibinfo{person}{Dayton~G Thorpe}, \bibinfo{person}{Andrew~J Duberstein}, {and} \bibinfo{person}{Ian~A Kinsey}.} \bibinfo{year}{2024}\natexlab{}.
\newblock \showarticletitle{Dubo-SQL: Diverse Retrieval-Augmented Generation and Fine Tuning for Text-to-SQL}.
\newblock \bibinfo{journal}{\emph{arXiv preprint arXiv:2404.12560}} (\bibinfo{year}{2024}).
\newblock


\bibitem[Wang et~al\mbox{.}(2023)]%
        {macsql}
\bibfield{author}{\bibinfo{person}{Bing Wang}, \bibinfo{person}{Changyu Ren}, \bibinfo{person}{Jian Yang}, \bibinfo{person}{Xinnian Liang}, \bibinfo{person}{Jiaqi Bai}, \bibinfo{person}{Qian-Wen Zhang}, \bibinfo{person}{Zhao Yan}, {and} \bibinfo{person}{Zhoujun Li}.} \bibinfo{year}{2023}\natexlab{}.
\newblock \showarticletitle{Mac-sql: Multi-agent collaboration for text-to-sql}.
\newblock \bibinfo{journal}{\emph{arXiv preprint arXiv:2312.11242}} (\bibinfo{year}{2023}).
\newblock


\bibitem[Wang et~al\mbox{.}(2019)]%
        {rat}
\bibfield{author}{\bibinfo{person}{Bailin Wang}, \bibinfo{person}{Richard Shin}, \bibinfo{person}{Xiaodong Liu}, \bibinfo{person}{Oleksandr Polozov}, {and} \bibinfo{person}{Matthew Richardson}.} \bibinfo{year}{2019}\natexlab{}.
\newblock \showarticletitle{Rat-sql: Relation-aware schema encoding and linking for text-to-sql parsers}.
\newblock \bibinfo{journal}{\emph{arXiv preprint arXiv:1911.04942}} (\bibinfo{year}{2019}).
\newblock


\bibitem[Wang et~al\mbox{.}(2017)]%
        {parse_survey2}
\bibfield{author}{\bibinfo{person}{Chenglong Wang}, \bibinfo{person}{Alvin Cheung}, {and} \bibinfo{person}{Rastislav Bodik}.} \bibinfo{year}{2017}\natexlab{}.
\newblock \showarticletitle{Synthesizing highly expressive SQL queries from input-output examples}. In \bibinfo{booktitle}{\emph{Proceedings of the 38th ACM SIGPLAN Conference on Programming Language Design and Implementation}}. \bibinfo{pages}{452--466}.
\newblock


\bibitem[Wang et~al\mbox{.}(2024)]%
        {llmagentsurvey}
\bibfield{author}{\bibinfo{person}{Lei Wang}, \bibinfo{person}{Chen Ma}, \bibinfo{person}{Xueyang Feng}, \bibinfo{person}{Zeyu Zhang}, \bibinfo{person}{Hao Yang}, \bibinfo{person}{Jingsen Zhang}, \bibinfo{person}{Zhiyuan Chen}, \bibinfo{person}{Jiakai Tang}, \bibinfo{person}{Xu Chen}, \bibinfo{person}{Yankai Lin}, {et~al\mbox{.}}} \bibinfo{year}{2024}\natexlab{}.
\newblock \showarticletitle{A survey on large language model based autonomous agents}.
\newblock \bibinfo{journal}{\emph{Frontiers of Computer Science}} \bibinfo{volume}{18}, \bibinfo{number}{6} (\bibinfo{year}{2024}), \bibinfo{pages}{1--26}.
\newblock


\bibitem[Wei et~al\mbox{.}(2023)]%
        {cot}
\bibfield{author}{\bibinfo{person}{Jason Wei}, \bibinfo{person}{Xuezhi Wang}, \bibinfo{person}{Dale Schuurmans}, \bibinfo{person}{Maarten Bosma}, \bibinfo{person}{Brian Ichter}, \bibinfo{person}{Fei Xia}, \bibinfo{person}{Ed Chi}, \bibinfo{person}{Quoc Le}, {and} \bibinfo{person}{Denny Zhou}.} \bibinfo{year}{2023}\natexlab{}.
\newblock \bibinfo{title}{Chain-of-Thought Prompting Elicits Reasoning in Large Language Models}.
\newblock
\newblock
\showeprint[arxiv]{2201.11903}~[cs.CL]
\urldef\tempurl%
\url{https://arxiv.org/abs/2201.11903}
\showURL{%
\tempurl}


\bibitem[Xiao et~al\mbox{.}(2023)]%
        {bge_embedding}
\bibfield{author}{\bibinfo{person}{Shitao Xiao}, \bibinfo{person}{Zheng Liu}, \bibinfo{person}{Peitian Zhang}, {and} \bibinfo{person}{Niklas Muennighoff}.} \bibinfo{year}{2023}\natexlab{}.
\newblock \bibinfo{title}{C-Pack: Packaged Resources To Advance General Chinese Embedding}.
\newblock
\newblock
\showeprint[arxiv]{2309.07597}~[cs.CL]


\bibitem[Xu et~al\mbox{.}(2017)]%
        {sqlnet}
\bibfield{author}{\bibinfo{person}{Xiaojun Xu}, \bibinfo{person}{Chang Liu}, {and} \bibinfo{person}{Dawn Song}.} \bibinfo{year}{2017}\natexlab{}.
\newblock \showarticletitle{Sqlnet: Generating structured queries from natural language without reinforcement learning}.
\newblock \bibinfo{journal}{\emph{arXiv preprint arXiv:1711.04436}} (\bibinfo{year}{2017}).
\newblock


\bibitem[Yu et~al\mbox{.}(2018a)]%
        {syntaxsqlnet}
\bibfield{author}{\bibinfo{person}{Tao Yu}, \bibinfo{person}{Michihiro Yasunaga}, \bibinfo{person}{Kai Yang}, \bibinfo{person}{Rui Zhang}, \bibinfo{person}{Dongxu Wang}, \bibinfo{person}{Zifan Li}, {and} \bibinfo{person}{Dragomir Radev}.} \bibinfo{year}{2018}\natexlab{a}.
\newblock \showarticletitle{Syntaxsqlnet: Syntax tree networks for complex and cross-domaintext-to-sql task}.
\newblock \bibinfo{journal}{\emph{arXiv preprint arXiv:1810.05237}} (\bibinfo{year}{2018}).
\newblock


\bibitem[Yu et~al\mbox{.}(2018b)]%
        {spider}
\bibfield{author}{\bibinfo{person}{Tao Yu}, \bibinfo{person}{Rui Zhang}, \bibinfo{person}{Kai Yang}, \bibinfo{person}{Michihiro Yasunaga}, \bibinfo{person}{Dongxu Wang}, \bibinfo{person}{Zifan Li}, \bibinfo{person}{James Ma}, \bibinfo{person}{Irene Li}, \bibinfo{person}{Qingning Yao}, \bibinfo{person}{Shanelle Roman}, {et~al\mbox{.}}} \bibinfo{year}{2018}\natexlab{b}.
\newblock \showarticletitle{Spider: A large-scale human-labeled dataset for complex and cross-domain semantic parsing and text-to-sql task}.
\newblock \bibinfo{journal}{\emph{arXiv preprint arXiv:1809.08887}} (\bibinfo{year}{2018}).
\newblock


\bibitem[Yu et~al\mbox{.}(2022)]%
        {modrag1}
\bibfield{author}{\bibinfo{person}{Wenhao Yu}, \bibinfo{person}{Dan Iter}, \bibinfo{person}{Shuohang Wang}, \bibinfo{person}{Yichong Xu}, \bibinfo{person}{Mingxuan Ju}, \bibinfo{person}{Soumya Sanyal}, \bibinfo{person}{Chenguang Zhu}, \bibinfo{person}{Michael Zeng}, {and} \bibinfo{person}{Meng Jiang}.} \bibinfo{year}{2022}\natexlab{}.
\newblock \showarticletitle{Generate rather than retrieve: Large language models are strong context generators}.
\newblock \bibinfo{journal}{\emph{arXiv preprint arXiv:2209.10063}} (\bibinfo{year}{2022}).
\newblock


\bibitem[Zhang et~al\mbox{.}(2023)]%
        {hallucination}
\bibfield{author}{\bibinfo{person}{Yue Zhang}, \bibinfo{person}{Yafu Li}, \bibinfo{person}{Leyang Cui}, \bibinfo{person}{Deng Cai}, \bibinfo{person}{Lemao Liu}, \bibinfo{person}{Tingchen Fu}, \bibinfo{person}{Xinting Huang}, \bibinfo{person}{Enbo Zhao}, \bibinfo{person}{Yu Zhang}, \bibinfo{person}{Yulong Chen}, {et~al\mbox{.}}} \bibinfo{year}{2023}\natexlab{}.
\newblock \showarticletitle{Siren's song in the AI ocean: a survey on hallucination in large language models}.
\newblock \bibinfo{journal}{\emph{arXiv preprint arXiv:2309.01219}} (\bibinfo{year}{2023}).
\newblock


\bibitem[Zhong et~al\mbox{.}(2020)]%
        {test-suite}
\bibfield{author}{\bibinfo{person}{Ruiqi Zhong}, \bibinfo{person}{Tao Yu}, {and} \bibinfo{person}{Dan Klein}.} \bibinfo{year}{2020}\natexlab{}.
\newblock \bibinfo{title}{Semantic Evaluation for Text-to-SQL with Distilled Test Suites}.
\newblock
\newblock
\showeprint[arxiv]{2010.02840}~[cs.CL]


\bibitem[Zhong et~al\mbox{.}(2017)]%
        {wikisql}
\bibfield{author}{\bibinfo{person}{Victor Zhong}, \bibinfo{person}{Caiming Xiong}, {and} \bibinfo{person}{Richard Socher}.} \bibinfo{year}{2017}\natexlab{}.
\newblock \showarticletitle{Seq2sql: Generating structured queries from natural language using reinforcement learning}.
\newblock \bibinfo{journal}{\emph{arXiv preprint arXiv:1709.00103}} (\bibinfo{year}{2017}).
\newblock


\end{thebibliography}

\end{document}